\documentclass[a4paper]{article} 

\usepackage{float}
\usepackage[
  a4paper,top=3cm,bottom=2cm,left=3cm,right=3cm,marginparwidth=1.75cm]{geometry}
\usepackage{amsmath}
\usepackage{calrsfs}
\usepackage{amsthm,bm,mathtools,etoolbox,amsfonts,bbm}
\usepackage{hyperref}
\hypersetup{colorlinks,linkcolor={blue},citecolor={blue},urlcolor={blue}}    
\usepackage[nameinlink,capitalise]{cleveref}
\usepackage[utf8]{inputenc} 
\usepackage[T1]{fontenc}    
\usepackage{url}          
\usepackage{booktabs}       
\usepackage{natbib}
\usepackage{placeins}
\usepackage{adjustbox}
\usepackage{booktabs}
\usepackage{algorithm}
\usepackage{algpseudocode}
\usepackage{multicol} 
\usepackage{makecell, multirow, tabularx} 
\usepackage{rotating}
\usepackage{float}
\usepackage{ulem}
\usepackage{graphicx}
\usepackage{tabularx}
\usepackage{authblk}
\usepackage{svg}

\newtheorem{Theorem}{Theorem}[section]

\newtheorem{Lemma}{Lemma}[section]
\newtheorem{Remark}{Remark}[section]
\newtheorem*{Remarknn}{Remark}
\newtheorem{Example}{Example}[section]

\title{Scalable Subsampling Inference for  Deep Neural Networks}

\author[1]{Kejin Wu}
\author[2]{Dimitris N. Politis}

\affil[1]{Department of Mathematics, University of California, San Diego }
\affil[2]{Department of Mathematics and Halicio\u{g}lu Data Science Institute, University of California, San Diego}

\date{}

\begin{document}

\maketitle

\begin{abstract}
Deep neural networks (DNN) has received increasing attention in machine learning applications in the last several years. Recently, a non-asymptotic error bound has been developed to measure the performance of the fully connected DNN estimator with ReLU activation functions for estimating regression models. The paper at hand gives a small improvement on the current error bound based on the latest results on the approximation ability of DNN. More importantly, however, a non-random subsampling technique--scalable subsampling--is applied to construct a `subagged'  DNN estimator. Under regularity conditions, it is shown that the subagged DNN estimator is computationally efficient without sacrificing accuracy for either estimation or prediction tasks. Beyond point estimation/prediction, we propose different approaches to build confidence and prediction intervals based on the subagged DNN estimator. In addition to being asymptotically valid, the proposed confidence/prediction intervals appear to work well in 
finite samples. All in all, the scalable subsampling DNN estimator offers the 
complete package in terms of statistical inference, i.e., 
(a) computational efficiency;
(b) point estimation/prediction accuracy; and
(c) allowing for the construction of practically useful 
confidence and prediction intervals.
\end{abstract}

\section{Introduction}\label{Sec:Intro}
In the last several years, machine learning (ML) methods have been developed rapidly fueled by ever-increasing amounts of data and computational power. Among different ML methods, a popular and widely-used technique is Neural Networks (NN) that models the relationship between inputs and outputs through layers of connected computational neurons. The idea of applying such a biology-analogous framework can be traced to the work of \cite{mcculloch1943logical}. 

At the end of the 20th century, people focused on the feed-forward Shallow Neural Networks (SNN) with sigmoid-type activation functions. An SNN has only one hidden layer but is shown to possess the universal approximation property, i.e., it can be used to approximate any Borel measurable function from one finite dimensional space to another with any desired degree of accuracy---see   \cite{cybenko1989approximation,hornik1989multilayer} and references within. However, the SNN practical performance left much to be desired. In the last ten or so years, Deep Neural Networks (DNN) received increased attention due to their great empirical performance.

Although DNN have become a state-of-the-art model, their theoretical foundation is still in development. Notably, \cite{yarotsky2018optimal,yarotsky2020phase} explored the approximation ability of DNN for any function $f$ that belongs to a H\"{o}lder Banach space; here, the sigmoid-type activation functions are now replaced by ReLU-type functions to avoid the gradient vanishing problem. The aforementioned work showed that the optimal  error
of the DNN estimator $f_{\text{DNN}} $ can be uniformly bounded,   %$W^{-2\xi/d}$ with the depth of the DNN, $L = O(W)$
 i.e., 
\begin{equation}\label{Eq: basicbound}
    || f - f_{\text{DNN}} ||_{\infty} =O\left(W^{-2\xi/d}\right);
\end{equation}
here, $\xi$ is some smoothness measurement of the target function $f: {\bf R}^d \to {\bf R} $
---see  \cref{Sec:EstDNN} for a formal definition; $W$ is the size of a neural network $f_{\text{DNN}}$, i.e., the total number of parameters; and $d$ is the dimension of the function inputs.

However, the bound (\ref{Eq: basicbound}) is not useful in practice. The reason is three-fold: (a) it requires a discontinuous weight assignment to build the desired DNN, so it is not feasible to train such DNN with usual gradient-based methods; (b) the structure of the DNN might not be the standard fully connected form so finding the satisfied specific structure becomes another difficult; most importantly, (c) this error bound is on the {\it optimal} estimation we can achieve from a finely designed DNN. It fails to tell us any story about the situation of applying the DNN estimator to solving real-world problems.

For example, what is the performance of the DNN to estimate a regression function with $n$ independent samples $\{(Y,\bm{X}_i)\}_{i=1}^{n}$ generated from an underlying true model $f$? It is easy to see that the error $\varepsilon$ of $f_{\text{DNN}}$ in sup-norm can be arbitrarily small if we allow $W$ to be arbitrarily large based on \cref{Eq: basicbound}. However, this optimal performance is hardly achievable and only represents the theoretically best estimation. What we attempt to do in this paper is to determine an empirically optimal $\widehat{f}_{\text{DNN}}$ with samples $\{(Y,\bm{X}_i)\}_{i=1}^{n}$ and then explore its estimation and prediction inference. Guided by this spirit, people usually think $\widehat{f}_{\text{DNN}}$ as an $M$-estimator and set different loss functions for various purposes:
\begin{equation}\label{Eq:widehatDNN}
\widehat{f}_{\text{DNN}} \in \arg\min_{f_{\theta}\in\mathcal{F}_{\text{DNN}}}\frac{1}{n}\sum_{i=1}^{n}L(f_{\theta}(\bm{x}_i),y_i);
\end{equation}
here $\mathcal{F}_{\text{DNN}}$ is a user-chosen space that contains all DNN candidates; $L(\cdot,\cdot)$ is the loss function, e.g., Mean Squared Errors loss for the regression problem with real-valued output, i.e., $L(f_{\theta}(\bm{x}_i),y_i) = (f_{\theta}(\bm{x}_i) - y_i)^2/2$; $\{(y,\bm{x}_i)\}_{i=1}^{n}$ are realizations of $\{(Y,\bm{X}_i)\}_{i=1}^{n}$.

In the paper at hand, we consider DNN-based estimation and prediction inference in the data-generating model: $Y_i = f(\bm{X}_i) + \epsilon_i$; 
here, the $\epsilon_i$ are independent, identically distributed  (i.i.d.) from a distribution $F_{\epsilon}$ that has mean 0 and variance $\sigma^2$---we will use the shorthand $\epsilon_i \sim$ i.i.d.~$(0,\sigma^2$). Consequently, $f(\bm{x}_i)=\mathbb{E}(Y_i|\bm{X}_i = \bm{x}_i)$. Furthermore, the regression function $f(\cdot)$ will be assumed to
 satisfy some smoothing condition which will be specified later. 
Note that the additive model with heteroscedastic error: $Y_i = f(\bm{X}_i) + g(\bm{X}_i)\cdot\epsilon_i$ can be analyzed similarly by applying two DNNs, one to estimate $f(\cdot)$ and one for $g(\cdot)$. 

From a nonparametric regression view, it is well-known that the optimal convergence rate of the estimation for a $p$-times continuously differentiable regression function of a $d$-dimensional argument is $n^{2p/(2p+d)}$---see \cite{stone1982optimal}.
If we assume the regression function belongs to a more general H\"{o}lder Banach space, 
we can define a non-integer $\xi = p + s$ to represent the smoothness order of $f$; here $0<s\leq 1$ is the H\"{o}lder coefficient. The optimal rate of non-parametric estimation can also be extended to such non-integer smoothness order; see Condition $3^{'}$ and  Definition 2 of \cite{kohler2023estimation}. Focusing on DNN estimation, the optimal and achievable error bound on the $L_2$ norm of $\widehat{f}_{\text{DNN}}$ is $O(n^{-\xi/(\xi+d)}\cdot \log^8(n))$ with a \textit{high probability}; this bound is due to  \cite{farrell2021deep} but the rate appears slower than the optimal rate that we can attain. Besides, although $\widehat{f}_{\text{DNN}}$ will become more accurate as the sample size increases, training DNN becomes very time-consuming. Moreover, it is infeasible to load massive data into a PC or even a supercomputer since its node memory is also limited in the computation process as pointed out by \cite{zou2021subbagging}. 

In this paper, we first give a small improvement on the bound of  \cite{farrell2021deep} using the latest results on the DNN approximation ability. Then, we resolve the computational issue involving huge data by applying the Scalable Subsampling technique of  \cite{politis2021scalable} to create a set of subsamples and then build a so-called subagging DNN estimator $\overline{f}_{\text{DNN}}$. Under regularity conditions, we can show that the subagging DNN estimator $\overline{f}_{\text{DNN}}$ could possess a faster convergence rate than a single DNN estimator $\widehat{f}_{\text{DNN}}$ trained on the whole sample. Lastly, using the same set of subsamples, we can build a Confidence Interval (CI) for $f$ based on $\overline{f}_{\text{DNN}}$. Due to the prevalent undercoverage phenomenon of CIs with finite samples, we propose two ideas to improve the empirical coverage rate: (1) we enlarge the CI by maximizing the margin of errors in an appropriate way; (2) we take an iterated subsampling method to build a specifically designed CI which is a combination of pivot-CI and quantile-CI. Beyond estimation inference, we also perform predictive inference (with both point and interval predictions).  

The paper is organized as follows. In \cref{Sec:DNN}, we give a short introduction to the structure of DNN. In \cref{Sec:SS}, we describe the methodology of scalable subsampling. Subsequently, the performance of the subagging DNN estimator and its associated confidence/prediction intervals are analyzed in \cref{Sec:EstDNN} and \cref{Sec:Preinf}. Simulations and conclusions are provided in \cref{Sec:Sim} and \cref{Sec:Conclu}, respectively. Proofs are given in \hyperref[Appendix:A]{Appendix: A} and additional simulations studies are presented in \hyperref[Appendix:B]{Appendix: B} and \hyperref[Appendix:C]{Appendix: C}.

In terms of notation, we will use the following norms: $\|g\|_{L_2(\boldsymbol{X})}: = \mathbb{E}[g(\boldsymbol{X})^2]^{1 / 2}$; $\|g\|_{\infty}: = \sup_{\bm{x}}|g(\bm{x})|$. Also, we employ  the notation $a_n=\Theta\left(d_n\right)$ to denote ``exact order'', i.e., that there exist two constants $c_1, c_2$ satisfying $c_1 \cdot c_2>0$, and $c_1 d_n \leq a_n \leq c_2 d_n$. We also use $\mathbb{E}_n[\cdot]$ to represent the sample average operator.

\section{Standard fully connected deep neural network}\label{Sec:DNN}
For completeness, we now give a brief introduction to the fully connected forward DNN with ReLU activation functions. Hereafter, we refer to DNN as the standard fully connected deep neural network with the ReLU activation function. In short, the DNN can be viewed as a parameterized family of functions. Its structure mainly depends on the input dimension $d$, depth $L\in \mathbb{N}$, width $\bm{H} \in \mathbb{N}^{L}$ and the output dimension. The depth $L$ describes how many hidden layers a DNN possesses; the width $\bm{H} = (H_1,\ldots,H_L)$ represents the number of neurons in each hidden layer. The forward property implies that the input, hidden neurons and output are connected in an acyclic relationship. The fully connected property indicates that each hidden neuron receives information from all hidden neurons at the previously hidden layer in a functional way. 

\begin{figure}[htbp]
 \centering
 \caption{The illustration of a fully connected DNN with $L = 2$, $H = 4$ and $W = 37$, and input dimension $d = 2$ and output dimension 1.}
 \bigskip 
 
 \includegraphics[scale = 0.35]{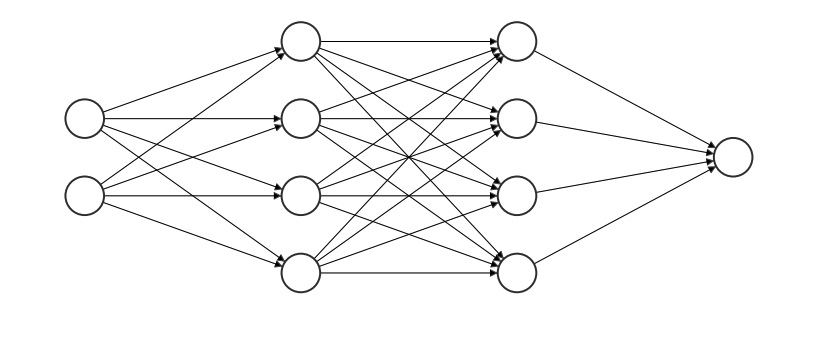}
 \label{FigDNNstructureex}
\end{figure}

Formally, if we let $\bm{u}_l = (u_{l,1},\ldots,u_{l,H_l})^{T}$ to represent all number of neurons at the $l$-th hidden layer for $l = 0, \ldots, L+1$; here, $\bm{u}_0$ represents the input vector $(x_1,\ldots,x_d)^{T}$ and $\bm{u}_{L+1}$ is the output. Therefore, we can pretend that the input layer and the output layer are the $0$-th and $(L+1)$-th hidden layers, respectively. Then, $u_{l,i} = \sigma(\bm{u}_{l-1}^{T}\bm{w}_{l-1,i} + b_{l-1,i})$ for $l = 1,\ldots,L$ and $i = 1,\ldots,H_l$; here $\bm{w}_{l-1,i}\in \mathbb{R}^{H_{l-1}}$ is the weight vector which connects the $(l-1)$-th hidden layer and the neuron $u_{l,i}$; $b_{l-1,i}\in \mathbb{R}$ is the corresponding intercept term; $\sigma(\cdot)$ is the so-called activation function and we take the ReLU function in this paper. To get the output layer, we just take $u_{L+1,i} = \bm{u}_{L}^{T}\bm{w}_{L,i} + b_{L,i}$ for $i = 1,\ldots,H_{L+1}$; here $H_{L+1}$ is equal to the output dimension. To express the functionality of the DNN in a more concise way, we can stack $\{\bm{w}_{l-1,i}^{T}\}_{i=1}^{H_{l}}$ by row to get $\bm{W}_{l-1}\in\mathbb{R}^{H_{l}}\times \mathbb{R}^{H_{l-1}}$ and collect $\{b_{l-1,i}\}_{i=1}^{H_l}$ to be a vector $\bm{b}_{l-1}$ for $l=1,\ldots, L+1$. Subsequently, we can treat the DNN as a function that takes the input $\bm{x}$ and returns output in the below way:
\begin{equation*}
    f_{\text{DNN}}(\bm{x}) =  \bm{W}_{L}(\sigma(\bm{W}_{L-1}(\cdots \sigma(\bm{W}_{2}\sigma(\bm{W}_{1}\sigma(\bm{W}_0\bm{x} + \bm{b}_{0}) + \bm{b}_1)  + \bm{b}_2)\cdots) +  \bm{b}_{L-1}) + \bm{b}_{L}.
\end{equation*}
We can understand that the function $f_{\text{DNN}}(\bm{x})$ maps $\bm{x}$ to the $1$-st hidden layer and then map the $1$-st hidden to the $2$-nd hidden layer and so on iteratively with weights $\{\bm{W}_l\}_{l=0}^{L}$, $\{\bm{b}_l\}_{l=0}^{L}$ and the activation function $\sigma(\cdot)$. We can then compute the total number of parameters in a DNN by the formula $W =  \sum_{i=0}^{L}(H_{i}\cdot H_{i+1} + H_{i+1})$. A simple DNN is presented in \cref{FigDNNstructureex}. It has a constant width of 4 and a depth of 2. 

\FloatBarrier

\section{Scalable subsampling}\label{Sec:SS}
Scalable subsampling is one type of non-stochastic \textit{subsampling} technique proposed by \cite{politis2021scalable}. Assume that we observe the sample $\{\bm{Z}_1, \ldots, \bm{Z}_n\}$; then, scalable subsampling relies on $q = \lfloor(n-b)/h \rfloor + 1$ number of
subsamples $B_1, \ldots, B_q$ where
$B_j = \{\bm{Z}_{(j-1)h+1}, \ldots,$ $\bm{Z}_{(j-1)h+b}\}$; here, $\lfloor \cdot \rfloor$ denotes the floor function, and $h$ controls the amount of overlap (or separation) between $B_j$ and $B_{j+1}$. In general, the subsample size $b$ and the overlap $h$ are functions of $n$, but these dependencies will not be explicitly denoted, e.g., 
$$ b = \Theta(n^{\beta})~;~h = a\cdot b,$$
%\footnote{let's just use one $\beta$, i.e., $b= \Theta\left(n^{\beta}\right)$  and $h\sim a \beta $ for some constant $a>0$}
where $0< \beta<1$ and $a>0$. More importantly, tuning $b$ and $h$ can make scalable subsampling samples have different properties. For example, if $h = 1$, the overlap is the maximum possible; if $h = 0.2b$, there is 80$\%$ overlap between $B_j$ and $B_{j+1}$; if $h = b$, there is no overlap between $B_j$ and $B_{j+1}$ but these two blocks are adjacent; if $h = 1.2 b$, there is a block of about $0.2b$ data points that separate the blocks $B_j$ and $B_{j+1}$. %Once we determine $b$ and $h$, we can build subagging estimator based on $\{B_j\}_{j=1}^{q}$. 

The \textit{bagging} idea was initially proposed by \cite{breiman1996bagging}, where the subsample is bootstrapped (sampling with replacement) with the same size as the original sample. As revealed by that work, the main benefit of taking this technique is that the mean-squared error (MSE) of the bagging estimator can decrease, especially for unstable estimators that may change a lot with different samples, e.g., neural networks and regression trees. There are ample works about combing the neural networks with the bagging technique to improve its generalization performance; e.g., see applications in the work of \cite{ha2005response,khwaja2015improved} for references. However, the drawback of the original bagging method is that the estimation process needs to be performed with $n$-size bootstrap resamples many times which is infeasible with massive data. 
\cite{buhlmann2002analyzing} proposed the \textit{subagging} idea which
is based on all subsamples as opposed to bootstrap resamples.  
However, even choosing a single random subsample could be computationally challenging when $n$ is large. As pointed out in \cite{ting2021simple}, drawing a random sample of size $b$ from $n$ items using the Sparse Fisher-Yates Sampler takes $O(b)$ time and space which corresponds to optimal time and space complexity for this problem. %If the bagging estimator is built with $T$ subsamples, the total computational cost of just generating random subsamples is of order $O(bT)$ which is   not tolerated  when both $b$ and $T$ are meant to tend to infinity.

%And then participators can create an aggregation estimator routinely based on all subsamples. People also try to combine the subsampling and bootstrap techniques to reduce the computational burden; see the BLB method from \cite{kleiner2014scalable} and SDB method from \cite{sengupta2016subsampled}. Nevertheless, these methods could still be computationally heavy due to the complexity of drawing subsamples being $O([an])$. 

Facing such computational dilemmas, scalable subsampling and 
subagging as proposed by \cite{politis2021scalable} can be seen as
an extension of the  Divide-and-Conquer principle---see e.g. \cite{jordan2013statistics}. Moreover, in addition to the computational savings, scalable subagging
may yield an estimator that is not less (and sometimes more) accurate 
than the original; the following example illustrates such a case.% technique is more computationally efficient for making estimation inferences about the subagging DNN estimator. To show the reasoning, we focus on the sample size to illustrate computational complexity and ignore other influential factors. 
%To describe this phenomenon, assume that the estimator can be computed in $O(n^{\zeta})$ operations for some constant $\zeta>0$, e.g., $\zeta = 1$ for sample mean or median of univariate data;  $\zeta = 2$ for sample mean of $d$-dimension real-vector data in which $d$ grows linearly with $n$. With the naive bagging method, the total number of operations to get the bagging estimator is $O(Tn^\zeta)$ when $T$ subsamples are generated. On the other hand, if we take $b=h$ for the scalable subsampling, the total number of operations to get the scalable subagging estimator is $O(qb^\zeta)$ which is $O(nb^{\zeta - 1})$ since $q = O(n/b)$. We should notice that the scalable subsampling is exempted from the computational cost of drawing subsamples, so it should still work more efficiently than the aforementioned standard subsampling, BLB and SDB methods even after simplifying the complexity of the latter two methods with weighted data representation. 

\begin{Example}[Kernel-smoothed function estimation]\label{Example: kernelestimator}
\normalfont
A remarkable example from the work of \cite{politis2021scalable} is the scalable subagging kernel estimator.  Suppose our goal is estimating the value of function $g$ at a specific point; here, the function $g$ can be a probability density, spectral density, or other function that is estimated in a nonparametric setting. Denote the estimand $\theta$ and its corresponding 
kernel-smoothed estimator $\hat{\theta}_n$ based on the whole sample,
and assume that   $\hat{\theta}_n$ satisfies the following conditions:
\begin{itemize}
    \item[(1)] $\mathbb{E}(\hat{\theta}_n^2)<\infty$ for all $n$;
    \item[(2)]  $n^\gamma(\mathbb{E}(\hat{\theta}_n)-\theta) \rightarrow C$ and $\text{Var}(n^\alpha\hat{\theta}_n) \rightarrow \sigma^2$ as $n \rightarrow \infty$, where $C$ is a non-zero constant, $\sigma^2>0$ and $\gamma>\alpha>0$.
\end{itemize}
Define the scalable subagging estimator as:
\begin{equation*}
    \bar{\theta}_{b, n, SS}=q^{-1} \sum_{i=1}^q \hat{\theta}_{b, i},
\end{equation*}
here $q$ is the total number of subsamples and $\hat{\theta}_{b, i}$ is the non-parametric estimator based on the $i$-th subsample $B_i$. To achieve the fastest convergence rate of $ \bar{\theta}_{b, n, SS}$
we may let $\beta = \frac{1}{1+2(\gamma-\alpha)}$. As a result, the Mean Squared Error (MSE )of the scalable subagging estimator $\bar{\theta}_{b,n,SS}$ is $\Theta(n^{-2 \gamma /(1+2(\gamma-\alpha))})$; see Lemma 4.1 of \cite{politis2021scalable} for a detailed discussion. To achieve such a convergence rate in the context of nonparametric estimation, the crucial point is using an undersmoothed bandwidth on the subsample statistics. %estimator so that the bias term is relatively negligible compared to the variance term; see \cite{politis2023multi} for more discussions about the effects of undersmoothing on the non-parametric estimations. Let's
To elaborate, suppose we are employing a non-negative (second-order) kernel for smoothing in which case the MSE-optimal bandwidth is $\Theta\left(n^{-1 / 5}\right)$. To conduct efficient subagging, however, the $\hat{\theta}_{b, i}$  should be computed using an undersmoothed bandwidth of order $o\left(b^{-1 / 5}\right)$. For example, if we choose the bandwidth for $\hat{\theta}_{b, i}$ to be $\Theta\left(b^{-1 / 4}\right)$ instead,
then the choices % yielding $\text{Bias}(\hat{\theta}_n)=O(n^{-1 / 2})$ and $\text{Var}(\hat{\theta}_n)=\Theta\left(n^{-3 / 4}\right)$; therefore, 
$\alpha=3 / 8$,  $\gamma=1 / 2$, $h=O(b)$, and 
$b= \Theta\left(n^{\beta}\right)$ with $\beta=0.8$ implies that the rate of convergence of $\bar{\theta}_{b,n,S S}$   is $n^{2 / 5}$. This rate is not only faster than the rate of $\hat{\theta}_n$ that used the sub-optimal bandwidth $\Theta\left(n^{-1 / 4}\right)$; it is actually the fastest rate achievable by any estimator that uses a non-negative kernel with its associated MSE-optimal bandwidth. Nevertheless, $\bar{\theta}_{b, n, S S}$ can be computed faster than $\hat{\theta}_n$, and may thus be preferable.
In addition to the asymptotic results, the simulation study of \cite{politis2021scalable} reveals that 
%on estimating probability density or spectral density functions
the error of the scalable subsampling estimator can actually be smaller than the
full-sample nonparametric estimator with its own optimal bandwidth choice. 
%This fact implies another advantage of the scalable subsampling technique, i.e., the bandwidth choice problem can be escaped as long as we apply the undersmoothing trick and then build a scalable subsampling estimator.  
 
\end{Example}

In the next section, we will introduce how to compute the scalable subsampling DNN estimator. 
%To achieve a possibly faster convergence rate, we still take $\beta_1 = \beta_2 = \beta$. 
%\footnote{please fix the way this assumption is motivated earlier....}
Then, we will show that our aggregated DNN estimator could possess a smaller MSE than the optimal DNN estimator trained on the whole sample, under some conditions. We also discuss some specifically designed confidence intervals to measure the estimation accuracy via the approaches mentioned in \cref{Sec:Intro}.

\section{Estimation inference with DNN}\label{Sec:EstDNN}
Although the DNN has captured much attention in practice, its theoretical validation is still in development. Recently, \cite{farrell2021deep} gave a non-asymptotic error bound to measure the performance of the DNN estimator under two regularity assumptions. To sync with the latest results on the approximation ability of DNNs, we replace their second assumption (Assumption 2 in their work) with the smooth function condition assumed by \cite{yarotsky2020phase}. We present these new assumptions below:
\begin{itemize}
\item A1: The regression data
are i.i.d.~copies of $\bm{Z}=(Y, \bm{X}) \in$ $\mathcal{Y} \times[-1,1]^d$, where $\bm{X}$ has a continuous distribution, and $\mathcal{Y} \subset[-M, M]$ for some positive constant $M$. Correspondingly, we set the space of all DNN candidate functions to be $ \mathcal{F}_{\text{DNN}} = \{ 
  f_{\theta}  : ||f_{\theta}||_{\infty}\leq 2M \}$.
\item A2: The target regression function $f$ lies in the H\"{o}lder Banach space $\mathcal{C}^{k, \alpha}\left([-1,1]^d\right)$ which is the space of $k$ times continuously differentiable functions on $[-1,1]^d$ having a finite norm
defined by
$$
\|f\|_{\mathcal{C}^{k, \alpha}\left([-1,1]^d\right)}=\max \left\{\max _{\mathbf{k}:|\mathbf{k}| \leq k} \max _{\mathbf{x} \in[-1,1]^d}\left|D^{\mathbf{k}} f(\mathbf{x})\right|, \max _{\mathbf{k}:|\mathbf{k}|=k} \sup _{\substack{\mathbf{x}, \mathbf{y} \in[-1,1]^d \\ \mathbf{x} \neq \mathbf{y}}} \frac{\left|D^{\mathbf{k}} f(\mathbf{x})-D^{\mathbf{k}} f(\mathbf{y})\right|}{\|\mathbf{x}-\mathbf{y}\|^\alpha}\right\},
$$
where the smoothness index is  $\xi=k+\alpha$ with an integer $k \geq 0$ and $0<\alpha \leq 1$. 

\item A3: The sample size $n$ is larger than $(2 e M)^2 \vee \text{Pdim}(f_{\text{DNN}})$ where $\text{Pdim}(f_{\text{DNN}})$ is the pseudo-dimension of $f_{\text{DNN}}$ which satisfies:
$$
c \cdot W L \log (W / L) \leq \operatorname{Pdim}(f_{\text{DNN}}) \leq C \cdot W L \log W,
$$
with some universal constants $c, C>0$ and Euler's number $e$; see \cite{bartlett2019nearly} for details.
\end{itemize}
\begin{Remarknn}
     We can weaken the assumption on the domain of $\boldsymbol{X}$ to $[-C_x,C_x]^{d}$ for some constant $C_x$, i.e., we can work on a compact domain of $\boldsymbol{X}$; see also the proof of \cite{yarotsky2020phase}.
% i.e, we first partition the whole space $[-C_x,C_x]^{d}$ into small patches and then approximate the function by DNN in each of those fine patches. 
\end{Remarknn}
As shown in \cite{farrell2021deep}, with $H_1 = H_2 = \cdots = H_L= \Theta(n^{\frac{d}{2(\xi+\alpha)}} \log ^2 n)$, $L = \Theta( \log n)$, the $L_2$ norm loss and empirical mean squared error of the deep fully connected ReLU network estimator from \cref{Eq:widehatDNN} can be bounded with probability at least $1-\exp \left(-n^{\frac{d}{\xi+d}} \log ^8 n\right)$, i.e.,
\begin{equation}\label{Eq: currentrate}
\begin{split}
     &\left\|\widehat{f}_{\text{DNN}}-f\right\|_{L_2(X)}^2 \leq C_1 \cdot\left\{n^{-\frac{\xi}{\xi+d}} \log ^8 n+\frac{\log \log n}{n}\right\}, \\
     &\mbox{and} \ \ \mathbb{E}_n\left[\left(\widehat{f}_{\text{DNN}}-f\right)^2\right] 
  = \Theta\left(\left\|\widehat{f}_{\text{DNN}}-f\right\|_{L_2(X)}^2\right);
\end{split}
\end{equation}
here $C_1>0$ is a constant which is independent of $n$ and may depend on $d, M$, and other fixed constants. 

Obviously, the $L_2$ norm error bound in (\ref{Eq: currentrate}) is sub-optimal compared to the fastest convergence rate we can achieve for nonparametric function estimation. 
With the latest approximation theory on DNNs, we can improve the error bound in \cref{Eq: currentrate} by decreasing the power of the $\log(n)$ term. Meanwhile, this faster rate is satisfied with a narrower DNN. We give our first theorem about the convergence rate of $\widehat{f}_{\text{DNN}}$ below. 

\begin{Theorem}\label{Theorem: 1}
Under assumptions A1 to A3, width $H = \Theta(n^{\frac{d}{2(\xi+\alpha)}} \log n)$, and depth $L = \Theta(  \log n$). Then, the $L_2$ norm loss of the deep fully connected ReLU network estimator \cref{Eq:widehatDNN} can be bounded with probability at least $1-\exp \left(-n^{\frac{d}{\xi+d}} \log ^6 n\right)$, i.e.,
\begin{equation}\label{Eq: latestrate}
    \left\|\widehat{f}_{\text{DNN}}-f\right\|_{L_2(X)}^2 \leq C_2 \cdot\left\{n^{-\frac{\xi}{\xi+d}} \log ^6 n+\frac{\log \log n}{n}\right\};
\end{equation}
here $C_2>0$ is another constant. 
\end{Theorem}
\noindent 
It appears that the above gives the fastest rate obtainable based on the current literature. Later, we will show how this error bound can be further improved by applying the scalable subagging technique under some mild conditions.  

\begin{Remarknn}
The improvement implied by \cref{Theorem: 1} can also be applied to  Corollaries 1 and 2 of \cite{farrell2021deep} to improve the corresponding error bounds.
\end{Remarknn}

\subsection{Scalable subagging DNN estimator}\label{Subsec: SSDNNest}
We first review the idea of scalable subagging and explain how this technique can be combined with DNN estimation. We focus on the regression problem and assume we observe sample $\{(Y_1,\bm{X}_1), \ldots, (Y_n, \bm{X}_n)\}$. 

Analogously to the subagging kernel-smoothed estimator of  \cref{Example: kernelestimator}, we can define the subagging DNN estimator as:
\begin{equation}\label{Eq: SSDNNest}
    \overline{f}_{\text{DNN}}(\bm{X}) = \frac{1}{q}\sum_{j=1}^q \widehat{f}_{\text{DNN},b,j}(\bm{X});
\end{equation}
here,  $q = \lfloor(n-b)/h \rfloor + 1$, and $\widehat{f}_{\text{DNN},b,j}(\cdot)$ is the minimizer of the empirical loss function in \cref{Eq:widehatDNN} just  using the data in the
$j$-th subsample namely $B_j = \{(Y_{(j-1)h+1}, \bm{X}_{(j-1)h+1}), \ldots,$ $(Y_{(j-1)h+b} ,\bm{X}_{(j-1)h+b})\}$.

%By \cref{Example: kernelestimator}, the key point of the scalable subagging is to perform estimation on non-random subsample sets with size $b$: $B_j = \{(Y_{(j-1)h+1}, \bm{X}_{(j-1)h+1}), \ldots,$ $(Y_{(j-1)h+b} ,\bm{X}_{(j-1)h+b})\}$ for $j =1,\ldots, q$; here, $h$ controls the amount of overlap (or separation) between $B_j$ and $B_{j+1}$ and $q = \lfloor(n-b)/h \rfloor + 1$ indicates the number of subsamples. When the bias for the estimator based on the subsample is tolerable, the subagging trick can yield a helpful variance deduction, especially for an unstable estimator which changes a lot with different data. 

In nonparametric function estimation where the estimation is performed through the kernel technique, the bandwidth can control the bias order of the kernel-smoothed estimator. As shown in \cref{Example: kernelestimator}, the optimal convergence rate can be recovered by combining scalable subagging trick and undersmoothing bandwidth. Similarly, in the context of neural network estimation, the whole architecture of a DNN controls its smoothness similar to the role of the shape (order) of a kernel. The depth and width of a DNN play the role of tuning parameters similar to the bandwidth of a kernel. Moreover, according to the prevailing wisdom, a deeper DNN may possess a lower bias;  this conjecture was confirmed by \cite{yang2020rethinking} with ResNet on some image datasets. 

However, as far as we know, there is no theoretical result that explains the relationship between bias and the width/depth of a DNN. Here, we make the below assumptions to restrict the order of the bias of $\widehat{f}_{\text{DNN}}$:
\begin{itemize}
    \item A4: $\mathbb{E}(\widehat{f}_{\text{DNN}}(\bm{x})  - f(\bm{x})) = O(n^{- \Lambda/2})$ uniformly in $\bm{x}$ for some constant $\Lambda >0$.
\end{itemize}  

%Here $\widehat{f}_{\text{DNN},b,i}(\cdot)$ is the minimizer of the empirical loss function in \cref{Eq:widehatDNN} on the $i$-th subsample.
To boost the scalable subagging method, a fundamental preliminary condition is that the bias of the estimator is comparatively negligible to its standard
deviation---see \cite{politis2021scalable} for details. Thus, we further impose an additional assumption on the order of bias:
\begin{itemize}
    \item A5: The bias exponent in Assumption A4 %order of $\widehat{f}_{\text{DNN}}$ 
satisfies the inequality: $\Lambda > \frac{\xi}{\xi + d}$. 
\end{itemize}
%\footnote{maybe combine A4 and A5 to one assumption?} 
 
We claim that assumptions A4 and A5 could be achievable. Due to the fact as revealed in \cite{yarotsky2020phase}, the approximation ability in the uniform norm of a DNN can be as fast as $W^{-2\xi/d}$. Although this rate is not instructive in practice, the existence of a DNN that satisfies the bias order requirement A4 is possible.

We should also notice that practitioners tend to build a large DNN whose size is larger than the sample size. i.e., the DNN interpolates the sample in the modern machine-learning practice. Interestingly, such an over-parameterized estimator breaks the classical understanding of the bias-variance trade-off since its generalization performance can even be better than a DNN which lies in the under-parameterized regime. Actually, this phenomenon is described as the double-descent of the risk by \cite{belkin2019reconciling}. Thus, A4 and A5 should be reasonable when we consider DNNs with an overwhelming number of parameters; however, assumption A3 may fail which means the consistency property of the DNN estimator may be lost. It is interesting to explore whether the scalable subsampling can work for DNN estimators in an over-parameterized regime; we leave this to future work. 

\begin{Remark}\label{Remark:complexityabalysis}
In this paper, we focus on applying scalable subagging to DNNs whose size is less than the sample size but the extension to a large DNN is straightforward. From the computability aspect, as we can expect, the saving of computational cost from applying scalable subagging will be more significant for executing estimation with a large DNN. To see this fact, let's assume that we consider a DNN with size $W = \Theta(n^{\phi})$, $\phi > 1$. Then, the computational complexity will be mainly determined by how many manipulations (e.g., forward calculation and backward updating) we carry out to train the DNN. The total number of manipulations is also affected by the batch size and the number of epochs. Thus, we summarize that the total number of manipulations is $O(n\cdot W \cdot E)$; here $E$ represents the number of epochs, i.e., the number of complete passes of the training through the algorithm. It is fair to assume that the complexity is in the order of $n^{\phi+1} := n^{\varphi}$. When the size of the DNN is larger than the sample size, $\varphi > 2$. Thus, for the subagging estimator, the computational complexity is approximately to be $O(n^{\beta\varphi}q) = O(n^{1+\beta(\varphi-1)})$. The ratio of $n^{1+\beta(\varphi-1)}$ over $n^{\varphi}$ is $n^{-(\varphi-1)(1-\beta)}$.
 Thus, for a fixed $\beta$, the larger $\varphi$ to be, the more computation can be saved by deploying the subagging technique. 
\end{Remark}
% Similarly with the subagging kernel estimator in \cref{Example: kernelestimator}, we can define the subagging DNN estimator as:
% \begin{equation}\label{Eq: SSDNNest}
%     \overline{f}_{\text{DNN}}(\bm{X}) = \frac{1}{q}\sum_{i=1}^q \widehat{f}_{\text{DNN},b,i}(\bm{X}).
% \end{equation}

Aggregating all the above,   the following theorem   quantifies the error bound of the scalable subagging DNN estimator of \cref{Eq: SSDNNest}:
\begin{Theorem}\label{Theorem: SSDNNerror}
Assume Assumptions A1 to A5,  and let $\beta = \frac{1}{1+ \Lambda - \frac{\xi}{\xi + d}}$.
Then, with   probability at least  $(1 - \exp(-n^{\frac{d}{\xi+d}}\log^6n))^q$ the error bound of the subagging estimator \cref{Eq: SSDNNest} in $L_2$ norm is:
\begin{equation*}
     \left\|\overline{f}_{\text{DNN}}-f\right\|_{L_2(X)}^2 \leq  n^{\frac{-\Lambda}{\Lambda + \frac{d}{\xi + d}}}  {\cal L}(n),
\end{equation*}
where ${\cal L}(n)$ is a slowly varying function involving a 
constant and all  $\log(n)$ terms.
\end{Theorem}
 
\begin{Remark}\label{Remark:notoptimalbeta}
 %   Notice that the choice $\beta = \frac{1}{1+ \Lambda - \frac{\xi}{\xi + d}}$ in \cref{Theorem: SSDNNerror} is not the optimal one to render the  smallest error bound since \footnote{what do you mean by "since"}  the variance order $\alpha$ is unknown. 
Choosing $\beta = \frac{1}{1+ \Lambda - \frac{\xi}{\xi + d}}$ in \cref{Theorem: SSDNNerror}  ensures that the square bias term will be always relatively negligible compared to the variance which is important for the
success of scalable subsampling; see related discussion in \cref{Remark:mu0}. 
\end{Remark}

%Since the order of the bias term is unclear at this moment, the error bound in \cref{Theorem: SSDNNerror} depends on $\Lambda$. We should also mention
 Note that the final accuracy of DNN heavily depends on many other factors in practice, e.g., which optimizer we choose in the training stage, which parameter initialization strategy we take, and how large the batch size should be. Thus, a solely theoretical rate is insufficient to verify the superiority of the scalable subsampling DNN estimator. We then deploy simulation studies in \cref{Sec:Sim} to provide supplementary evidence. 

\subsection{Estimation of the bias order of DNN estimator}\label{Susec:estbiasorder}
Although \cref{Theorem: SSDNNerror} shows the possibility of getting a smaller error bound, it depends on the bias exponent $\Lambda$ which is typically
unknown. In this subsection, we propose two approaches to estimate the value of $\Lambda$ via subsampling. As far as we know, it is the first attempt
towards quantifying the bias of the DNN estimator.

First note that A4 implies that, for any $i$, 
  $\mathbb{E}(\widehat{f}_{\text{DNN},b,i}(\bm{x})  - f(\bm{x})) = O(n^{- \beta\Lambda/2})$. 
Since $  \overline{f}_{\text{DNN}}(\bm{X}) = \frac{1}{q}\sum_{j=1}^q \widehat{f}_{\text{DNN},b,j}(\bm{X})$, it follows that
the  bias of $\overline{f}_{\text{DNN}}(\bm{x})$ is $ O(n^{- \beta\Lambda/2})$, so we can write
\begin{equation}\label{Eq:bias_c_b}
\mathbb{E}(\overline{f}_{\text{DNN}}(\bm{x}) - f(\bm{x})) = c_b\cdot b^{-\Lambda/2} + o(b^{-\Lambda/2}).
\end{equation}
Recall that $\overline{f}_{\text{DNN}}(\bm{X})$ was built based on   subsamples of size $b$. 
 If we have another DNN estimator $\widehat{f}_{\text{DNN},b_0}(\bm{x})$ trained on   sample of  size $b_0$, then its bias will be $c_b\cdot b_{0}^{-\Lambda/2} + o(b_{0}^{-\Lambda/2})$. Then, 
\begin{equation}\label{Eq:biasest}
\begin{split}
\mathbb{E}\left(\overline{f}_{\text{DNN}}(\bm{x}) - f(\bm{x})\right)  &= \mathbb{E}\left(\overline{f}_{\text{DNN}}(\bm{x}) - \widehat{f}_{\text{DNN},b_0}(\bm{x}) + \widehat{f}_{\text{DNN},b_0}(\bm{x}) - f(\bm{x})\right)\\
& =  \mathbb{E}\left(\overline{f}_{\text{DNN}}(\bm{x}) - \widehat{f}_{\text{DNN},b_0}(\bm{x})\right) + \mathbb{E}\left( \widehat{f}_{\text{DNN},b_0}(\bm{x}) - f(\bm{x})\right).
\end{split}
\end{equation}
If $b\to\infty$ and $b/b_0\to 0$, the bias of $\overline{f}_{\text{DNN}}(\bm{x})$ is asymptotically determined by the first term on the r.h.s. of \cref{Eq:biasest}. So we can try to estimate $\mathbb{E}\left(\overline{f}_{\text{DNN}}(\bm{x}) - \widehat{f}_{\text{DNN},b_0}(\bm{x})\right)$ to approximate the  l.h.s. of \cref{Eq:biasest}. 

Ideally, if we have a large enough sample, we can carve out $M$  
non-overlapping  (or partially
overlapping) $b_0$-size subsamples and compute $\{\widehat{f}_{\text{DNN},b_0}^{(i)}(\bm{x})\}_{i=1}^{M}$. If we further separate each $b_0$-size subsample into multiple non-overlapping (or partially
overlapping) $b$-size subsamples, $\{\overline{f}_{\text{DNN}}^{(i)}(\bm{x})\}_{i=1}^{M}$ can be built and each $\overline{f}_{\text{DNN}}^{(i)}(\bm{x})$ possesses the same bias order as our desired DNN estimator. Subsequently, the bias of $\overline{f}_{\text{DNN}}(\bm{x})$ can be estimated by the sample mean of $\{\overline{f}_{\text{DNN}}^{(i)}(\bm{x})-\widehat{f}_{\text{DNN},b_0}^{(i)}(\bm{x})\}_{i=1}^{M}$. We can then
use this information to estimate the value of $\Lambda$. By the law of large numbers, we can get accurate bias estimation as $M\to\infty$. However, as we can easily see, this approach is computationally heavy and requires a large dataset. 

Consequently, we propose another way to perform the bias estimation; we will call it \textit{scaling-down} estimation method. 
To elaborate, recall that our goal is estimating the bias of $\overline{f}_{\text{DNN}}(\bm{x})$ that was built based on subsamples of size $b$. 
 Consider different DNN estimators $\widehat{f}_{\text{DNN},b_1}(\bm{x})$ and $\widehat{f}_{\text{DNN},b_2}(\bm{x})$ which are trained 
on samples of size   $b_1$  and $b_2$ respectively; here $b_1 \ll b$ and $b_2 \ll b_1$. %And then we take a \textit{scaling-down} trick to estimate the bias of $\overline{f}_{\text{DNN}}(\bm{x})$. Similarly with the previous analysis, we assume
As before, A4 implies that the bias of $\widehat{f}_{\text{DNN},b_i}(\bm{x})$ is $c_b\cdot b_{i}^{-\Lambda/2} + o(b_{i}^{-\Lambda/2})$ for $i = 1, 2$. Then, a key observation is that:
\begin{equation}\label{Eq:biasest_scale}
\begin{split}
\mathbb{E}\left(\widehat{f}_{\text{DNN},b_i}(\bm{x})- f(\bm{x})\right)  &= \mathbb{E}\left(\widehat{f}_{\text{DNN},b_i}(\bm{x}) - \overline{f}_{\text{DNN}}(\bm{x}) + \overline{f}_{\text{DNN}}(\bm{x}) - f(\bm{x})\right)\\
& =  \mathbb{E}\left(\widehat{f}_{\text{DNN},b_i}(\bm{x}) - \overline{f}_{\text{DNN}}(\bm{x})\right) + \mathbb{E}\left( \overline{f}_{\text{DNN}}(\bm{x}) - f(\bm{x})\right),~\text{for}~ i = 1, 2.
\end{split}
\end{equation}
Due to the relationship between $b, b_1, b_2$, the bias of $\widehat{f}_{\text{DNN},b_i}(\bm{x})$ is dominated by the first term on the r.h.s. of \cref{Eq:biasest_scale}. We   then  have two different estimates of the bias of $\widehat{f}_{\text{DNN},b_i}(\bm{x})$, namely:
$$
\widehat{B}_i =  \frac{1}{q_i}\sum_{j=1}^{q_i}\left( \widehat{f}_{\text{DNN},b_i}^{(j)}(\bm{x}) -  \overline{f}_{\text{DNN}}(\bm{x}) \right),~\text{for}~ i = 1, 2.
$$
Fixing the value of $i$, 
$\{\widehat{f}_{\text{DNN},b_i}^{(j)}(\bm{x})\}_{j=1}^{q_i}$ is  
  value  of $\widehat{f}_{\text{DNN},b_i}(\bm{x})$ 
computed from the $j$th subsample of size $b_i$
carved out the whole sample; as before, these subsamples can
be non-overlapping or partially overlapping and their number is
denoted by $q_i$. Ignoring the $o(\cdot)$ term
in \cref{Eq:bias_c_b}, we can solve the following system of equations  to approximate both $c_b$ and $\Lambda$:
\begin{equation}\label{Eq:linearSys}
    \begin{cases} 
    \widehat{B}_1  &=  c_b\cdot b_1^{-\Lambda/2}  \\
    \widehat{B}_2  &=  c_b\cdot b_2^{-\Lambda/2}. \\
\end{cases}
\end{equation}
Taking logarithms in \cref{Eq:linearSys} turns it into
a linear system in $c_b$ and $\Lambda$.
 Finally, we can estimate the bias of $\overline{f}_{\text{DNN}}(\bm{x})$ by {\it scaling down} $\widehat{B}_1$ by a factor $(b/b_1)^{-\Lambda/2}$, i.e., the bias of $\overline{f}_{\text{DNN}}(\bm{x})$ is
approximately $\widehat{B}_1\cdot (b/b_1)^{-\Lambda/2}$. We summarize this procedure in \cref{Algo: Biasest}.

\begin{algorithm}[htbp]
\caption{\textit{Scaling-down} bias estimation of DNN estimator} 
%\centering
\label{Algo: Biasest}
  %\centering
  \begin{tabularx}{\textwidth}{lX}   
    Step 1 & Fix a subsample size $b$,  and compute $\overline{f}_{\text{DNN}}(\bm{x})$ at point $\bm{x}$. \\
    Step 2 & Fix two subsample sizes $b_1 \ll b$ and $b_2 \ll b_1$, 
and separate the  whole sample into $q_1$ and $q_2$ number of $b_1$-size and $b_2$-size subsamples, respectively. Compute $\{\widehat{f}_{\text{DNN},b_i}^{(j)}(\bm{x})\}_{j=1}^{q_i}$ at $\bm{x}$ for $i = 1, 2$.\\
    Step 3 & Solve \cref{Eq:linearSys} to get $c_b$ and $\Lambda$. \\
    Step 4 & Estimate the bias of $\overline{f}_{\text{DNN}}(\bm{x})$ by $\widehat{B}_1\cdot (b/b_1)^{-\Lambda/2}$.
  \end{tabularx}
\end{algorithm}

\subsection{Confidence intervals}
Beyond  point  estimation, it is important to quantify DNN estimation  accuracy; 
this can be done via a standard error or --even better-- via a Confidence Interval (CI).  More specifically, for a  point
of interest $\bm{X} = \bm{x}$, we hope to find a CI which satisfies:
\begin{equation*}
    \mathbb{P}(B_{l}\leq f(\bm{x}) \leq B_{u} ) =  1 - \delta;
\end{equation*}
here $\mathbb{P}$ should be understood as the conditional probability given $\bm{X} = \bm{x}$; $B_{l}$ and $B_{u}$ are lower and upper bound for $f(\bm{x})$
that are functions of the DNN estimator; $\delta$ is the significance level. Since 
we can have different CI constructions having the same $\delta$,  we are also interested in the CI length (CIL) which is defined as $\text{CIL} =B_{u} - B_{l} $. We aim for a (conditional) CI that is the most accurate
(in terms of its coverage being close to $1 - \delta$) but with the shortest length. 

Analogously to \cref{Example: kernelestimator}, we make an assumption about the variance term of the DNN estimator trained with  sample size $n$ and evaluated at $\bm{x}$:
\begin{itemize}
    \item [B1] Var$(n^\alpha \widehat{f}_{\text{DNN}}(\bm{x}))\to \sigma^2 >0$ as $n\to\infty$.
\end{itemize}
Generally speaking, we have two choices to build CI for $f(\bm{x})$: (1) Pivot-CI (PCI), the type of CI obtained by estimating the 
sampling distribution of a pivotal quantity, e.g. the estimator centered at its expectation;  (2) Quantile-CI (QPI), the type of CI based on quantiles of the estimated sampling distribution of the (uncentered) estimator of interest. More details are given in the example below.

\begin{Example}[Types of CI]\label{Example: typeCI}
\normalfont
For any unknown quantity $\theta$ estimated by
$\hat \theta_n$,  we may build a scalable subagging estimator $\bar{\theta}_{b, n, SS}=q^{-1} \sum_{i=1}^q \hat{\theta}_{b, i}$ to approximate it. To
construct a CI for $\theta$ based on $\bar{\theta}_{b, n, SS}$, we are aided 
by the CLT of \cite{politis2021scalable}, i.e., 
\begin{equation}\label{Eq:limitdistri}
   \kappa_n(\bar{\theta}_{b, n, SS} - \theta) \overset{d}{\to} N(C_{\mu},C_\sigma^2),~\text{as}~n \to \infty,
\end{equation}
under mild conditions; here $C_{\mu}$ and $C_\sigma^2$ are the mean and variance of limiting distribution, respectively, and $\kappa_n=n^{\frac{1-\beta+2\alpha\beta}{2}}$. [By the way, note the typo in
\cite{politis2021scalable} where $\kappa_n
$ was  incorrectly written as $n^{-\frac{1-\beta+2\alpha\beta}{2}}$.]

% As mentioned above, there are   two general classes of CIs. For PCI, we need to capture the limiting distribution of $\kappa_n(\bar{\theta}_{b, n, SS} - \theta)$, here $\kappa_n$ is a function of whole sample size $n$; the choice of $\kappa_n$ in the context of DNN estimators will be discussed later. We denote $J_n(z) = \mathbb{P}(\kappa_n(\bar{\theta}_{b, n, SS} - \theta)\leq z )$ and $J(z)$ is the limiting distribution of $J_{n}(z)$. Usually, due to the CLT for triangular arrays, see Theorem B.0.1 of \cite{politis1999subsampling} for help, $J(z)$ is the CDF of a normal distribution under some mild conditions, i.e.,

%To apply the above CLT for triangular arrays, we need to check three conditions of Theorem B.0.1 in \cite{politis1999subsampling}. For the first and third ones, they are satisfied due to the assumption imposed on $\mathcal{F}_{\text{DNN}}$ and the independence of different DNN estimators on subsamples in the context of this paper. Thus, 

The form of the PCI based on CLT (\ref{Eq:limitdistri}) depends crucially on 
whether $C_{\mu}=0$ or not; see the next two subsections for details.
On the other hand, the QCI is easier to build but it has its own
deficiencies. %more convenient to be built since we do not need to compute the margin-of-error term explicitly and we even do not need to worry about the bias issue. 
In the context of this example, it is tempting to create a QCI for $\theta$ by taking the $\delta/2$ and $1-\delta/2$ quantile values of the empirical distribution of the points $\{\hat{\theta}_{b,1},\ldots,\hat{\theta}_{b,q}\}$. However, the resulting CI will be too conservative, i.e., its coverage will be (much) bigger than
$1-\delta$. % compared to PCI built based on the same sample set. 
The reason is that the empirical distribution of 
$\{\hat{\theta}_{b,i},\ldots,\hat{\theta}_{b,q}\}$ is approximating the sampling distribution of estimator
$\hat{\theta}_{b}$ which has bigger variance than that of the
target $\hat{\theta}_{n}$. We could try to re-scale the  empirical 
distribution of $\{\hat{\theta}_{b,1},\ldots,\hat{\theta}_{b,q}\}$
as in classical subsampling---see \cite{politis1999subsampling}.
% However, in the context of DNN estimation, it is infeasible to do rescaling since the exact order $\alpha$ is unknown, {\color{cyan} but we can estimate it as the same method we used to estimate the bias order in \cref{Susec:estbiasorder}}. 
% \footnote{we can do subsampling with estimated rate as in 
% Ch 9 of \cite{politis1999subsampling};
% please add a Subsection ``Estimation of the variance order of DNN"
% and maybe add simulations? Here is the idea:
% Take two subsample sizes $b_1$ and $b_2$ as in Section 4.2, and 
% create $\{\hat{\theta}_{b_i,1},\ldots,\hat{\theta}_{b_i,q_i}\}$ for i=$1,2$.
% Let $V_i$ be the sample variance of $\{\hat{\theta}_{b_i,1},\ldots,\hat{\theta}_{b_i,q_i}\}$. 
% Then, by assumption B1: $V_i \approx b_i^{-2\alpha} \sigma^2$ for i=$1,2$.
% These are two equations in two unknowns $\alpha, \sigma^2$;
% take logs to solve it, estimate $ \alpha $ and re-scale the QCI
% }
We still consider the QCI in the simulation studies. As expected, this QCI is the most conservative one; see details in \cref{Sec:Sim}.
\end{Example}

\subsubsection{PCI in the case where  $C_{\mu} = 0$}
\label{Subsec:mu0}
If $C_{\mu} = 0$, i.e., when the square bias is relatively negligible compared to the variance in estimation,   we can rely on \cref{Eq:limitdistri} to build a PCI for the true function $f$ at a point $\bm{x}$.
All  we need is a consistent estimator of $C_\sigma^2$, e.g., $\widehat{C}_{\sigma}^2 = b^{2 \alpha} q^{-1} \sum_{i=1}^q\left(\hat{\theta}_{b, i}-\bar{\theta}_{b, n, S S}\right)^2$. In that case, a PCI for $\theta$ based on the CLT can be written as: 
\begin{equation}\label{Eq:PCInaive}
\bar{\theta}_{b, n, SS} \pm z_{1-\delta/2}\cdot\widehat{C}_{\sigma}\cdot\kappa_n^{-1},
\end{equation}
where $z_{1-\delta/2}$ is the  $1-\delta/2$ quantile of the standard normal distribution.
% However,   difficulty arises when the mean of the limiting distribution is non-zero, i.e., $C_{\mu} \neq 0 $. In this case, we need to do some debiasing manipulations, otherwise, the  CI   \cref{Eq:PCInaive} would be incorrect;
 
 Observing that there is a common term $n^{\beta\alpha}$ in $\kappa_n$ and $\widehat{C}_{\sigma}$, we can estimate $\widehat{C}_{\sigma}\cdot\kappa_n^{-1}$ as a whole  rather than computing $\kappa_n$ and $\widehat{C}^2_{\sigma}$ separately. As a result, we can get a simplified PCI based on \cref{Eq:PCInaive} as follows: 

\begin{equation}\label{Eq:simplePCI}
\overline{f}_{\text{DNN}}(\bm{x}) \pm z_{1-\delta/2}\cdot M_{\sigma}; 
\end{equation}
here $M_{\sigma}=\widehat{C}_{\sigma}\cdot\kappa_n^{-1}$ which can be approximated by $\sqrt{q^{-1} \sum_{i=1}^q\left(\widehat{f}_{\text{DNN},b,i}(\bm{x})- \overline{f}_{\text{DNN}}(\bm{x}) \right)^2}/n^{\frac{1-\beta}{2}}$. Note that the building of the CI does not require the knowledge of $\alpha$ which is the order of the variance term in B1.
%\footnote{but we could also estimate $\alpha$ as in the previous footnote}
However, the estimation $\widehat{C}_\sigma$ may not be accurate when $q$ is small since it is only an average of $q$ terms. %\footnote{why?}     
As a result, the PCI according to \cref{Eq:simplePCI} may undercover the true model values. Thus, we may relax the desired property of CI. Instead of requiring the exact coverage rate of a CI to be $1-\delta$, we seek a CI such that:
\begin{equation}\label{Eq:finalCIgoal}
    \mathbb{P}(B_{l}\leq f(\bm{x}) \leq B_{u} ) \geq 1 - \delta.
\end{equation}
Thus, the optimal candidate will be the CI which has the shortest length and guarantees the lowest coverage rate larger than $1-\delta$. To satisfy \cref{Eq:finalCIgoal}, we  may enlarge the CI appropriately by replacing $\widehat{C}^2_{\sigma}$ with $\widetilde{C}^2_{\sigma} = \widehat{C}^2_{\sigma} + (\overline{f}_{\text{DNN}}(\bm{x}) - y)^2$; here $y = f(\bm{x}) + \epsilon$.

It is appealing to think that $\widetilde{C}^2_{\sigma}$ is close to the MSE of $\overline{f}_{\text{DNN}}(\bm{x})$. However,  
$$ (\overline{f}_{\text{DNN}}(\bm{x}) - y)^2 = \left(\frac{1}{q}\sum_{i=1}^{q}(\widehat{f}_{\text{DNN},b,i}(\bm{x}) - y) \right)^2 = \left(\frac{1}{q}\sum_{i=1}^{q}\left(\widehat{f}_{\text{DNN},b,i}(\bm{x}) - f(\bm{x})\right)-\epsilon \right)^2.$$
When $q$ is large, $(\overline{f}_{\text{DNN}}(\bm{x}) - y)^2\to(C_{\mu} - \epsilon)^2$ where $C_{\mu}$ is the bias of $\overline{f}_{\text{DNN}}(\bm{x})$. Therefore, $\widetilde{C}^2_{\sigma}$ is not exactly the MSE, but it can still be used to enlarge the CI to some extent. We can then define another PCI as:
\begin{equation}\label{Eq:simplePCIenlarged}
\overline{f}_{\text{DNN}}(\bm{x}) \pm z_{1-\delta/2}\cdot \widetilde{M}_{\sigma},
\end{equation}
where  
\begin{equation}\label{Eq:twoenlargedPCI}
\widetilde{M}_{\sigma} = \sqrt{ q^{-1} \sum_{i=1}^q\left(\widehat{f}_{\text{DNN},b,i}(\bm{x})- \overline{f}_{\text{DNN}}(\bm{x})\right)^2/n^{1-\beta} + (\overline{f}_{\text{DNN}}(\bm{x}) - y)^2/n^{1-\beta+2\alpha\beta}}.
\end{equation}
\noindent 
Since the order of the variance term $\alpha$ is involved in the above terms, we consider two extreme situations in the simulation sections: (1) We take $2\alpha = 0 $ which is a most enlarged case; or (2)  take $2\alpha = 1$ which is a mildly enlarged case. 
%\footnote{maybe estimate $\alpha$?}

\begin{Remark}[The condition to guarantee $C_{\mu} = 0$]\label{Remark:mu0}
    According to \cref{Eq:limitdistri}, $C_{\mu} = 0$ is satisfied as long as $\beta >\frac{1}{1+\Lambda - 2\alpha}$ under A5. If we take $\beta =  \frac{1}{1+ \Lambda - \frac{\xi}{\xi + d}}$ in \cref{Theorem: SSDNNerror}, we can find that the condition for $C_{\mu} = 0$ is always satisfied. This is not surprising due to A5 imposing the requirement on the convergence rate of the bias term. However, as explained in \cref{Remark:notoptimalbeta}, this $\beta$ is not the optimal one to generate the smallest error bound. Thus, we could arrive at a stage where the orders of squared bias and variance are the same once we know $\alpha$. Due to the high variability of training a DNN in practice, we introduce a method in \cref{Subsec:munot0} to build CI appropriately under the situation that $C_{\mu} \neq 0$, which serves for cases where the bias is not relatively negligible.
\end{Remark}

\subsubsection{PCI in the case where  $C_{\mu} \neq 0$}\label{Subsec:munot0}
It is worthwhile to discuss how can we build a PCI for scalable subsampling DNN estimator when $C_{\mu} \neq 0$. Note that \cite{politis2021scalable} proposed an {\it iterated} scalable subsampling technique that is applicable in the case $C_{\mu} \neq 0$. While this technique is also applicable in the case $C_{\mu} = 0$, we may prefer the construction of \cref{Subsec:mu0} since it is less computer-intensive. However, we should notice that the additional computational burden brought by {\it iterated} subsampling is negligible when $n\to\infty$; see analysis in \hyperref[Appendix:C]{Appendix: C}. For completeness, we present this method here in the remark below. 

\begin{Remark}[Iterated subsampling] \label{Remark:iteratedsubsampling}
With the same notations in \cref{Example: typeCI}, we can perform the iterated subsampling in three steps: (1) Let $b=\left\lfloor n^\beta\right\rfloor$, then apply the scalable subsampling technique to sample $X_1, \ldots, X_n$ and get $q$ subsets $\{B_i\}_{i=1}^q$. Compute $\bar{\theta}_{b, n, SS}$; we call it ``first stage subsampling''; (2) Take another subsample size $b^{\prime} = \lfloor b^{\beta} \rfloor$  and apply scalable subagging method again to all $\{B_i\}_{i=1}^q$, i.e., as if $B_i$ where the only data at hand and make subagging estimator for each $B_i$ subsamples; such subagging estimator $\bar{\theta}_{b^{\prime}, b, SS,i}$ is computed by averaging $q^{\prime}$ estimators $\{\hat{\theta}_{b^{\prime}, b, SS,i}^{(j)}\}_{j=1}^{q^{\prime}}$; here $q^{\prime} = \lfloor(b-b^{\prime})/h^{\prime} \rfloor + 1$. As a result, we can get $q$ number of $\{\bar{\theta}_{b^{\prime}, b, SS,i}\}_{i=1}^{q}$; we call it ``iterated stage subsampling''; (3) Find the subsampling distribution $L_{b^{\prime}, b, S S}(z)=$ $q^{-1} \sum_{i=1}^q 1\left\{\kappa_b\left(\bar{\theta}_{b^{\prime}, b, SS,i}-\bar{\theta}_{b, n, S S}\right) \leq z\right\}$; $\kappa_b$ is a function of $b$. In the context of DNN estimation, we use $\widehat{f}^{(j)}_{\text{DNN},b,i}$ to represent the DNN estimator in the iterated subsampling stage on the $j$-th subsamples from the $i$-th subsample in the first stage subsampling. 

%In \hyperref[Appendix:C]{Appendix: C}, we show that the computational burden of iterated subsampling is relatively negligible compared to the first stage subsampling when $n$ is large. 
%\footnote{what is "first stage subsampling"}
\end{Remark}

 Denote $J_n(z) = \mathbb{P}(\kappa_n(\bar{\theta}_{b, n, SS} - \theta)\leq z )$, and $J(z)$ is the limit of $J_{n}(z)$ as $n\to \infty$; recall that 
(\ref{Eq:limitdistri}) implied that $J(z)$ is Gaussian.
 Proposition 2.1 of \cite{politis2021scalable} shows that $L_{b^{\prime}, b, S S}(z)$ converges to $J(z)$ in probability for all points of continuity of $J(z)$. Due to \cref{Eq:limitdistri}, $J(z)$ is continuous everywhere, and therefore the
convergence is uniform. Thus, both $L_{b^{\prime}, b, S S}(z)$ and $J_n(z)$ converge in a uniform fashion to $J(z)$ in probability  which implies   that:
\begin{equation}
\sup _z\left|L_{b^{\prime}, b, S S}(z) - J_n(z)\right| \overset{p}{\to} 0,~\text{as}~n\to\infty.
\end{equation}
Thus,   iterated subsampling can be used to estimate the distribution $J_n$. We can build the CI in a pivotal style without explicitly referring to the form of 
$J$ that involves the two unknown parameters. A further
 issue is that   normality might not be 
well represented in $J_n$   since it is based on an average of $q$ 
quantities; having a large $q$ requires a huge $n$. To compensate for the data size requirement, we take a specific approach to build CI which can be considered as a combination of PCI and QCI to some extent. 
\cref{Algo: PCI} describes all the steps to construct the CI for $f$ at a point $\bm{x}$ based on the subagging DNN estimator and iterated subsampling method.

\begin{algorithm}[htbp]
\caption{PCI of $f(\bm{x})$ based on iterated subsampling} 
%\centering
\label{Algo: PCI}
  %\centering
  \begin{tabularx}{\textwidth}{lX}   
    Step 1 & Fix the subsample size $b$, compute $\overline{f}_{\text{DNN}}(\bm{x})$ at point $\bm{x}$. \\
    Step 2 & Fix the subsample size $b^{\prime}$ of iterated subsampling, perform necessary steps in \cref{Remark:iteratedsubsampling} to find 
    $$
    L_{b^{\prime}, b, S S}(z)= q^{-1} \sum_{i=1}^q 1\left\{\kappa_b\left(\overline{f}_{\text{DNN},i}(\bm{x})-\overline{f}_{\text{DNN}}(\bm{x}) \right) \leq z\right\};
    $$ 
    here $\overline{f}_{\text{DNN},i}(\bm{x}) = \frac{1}{q^{\prime}}\sum_{j=1}^{q^{\prime}}\widehat{f}^{(j)}_{\text{DNN},b,i}$ is the subagging DNN estimator on the $i$-th subsamples in Step 1 at the point $\bm{x}$. 
    \\
    Step 3 & Denote the $\delta/2$ and $1-\delta/2$ quantile values of the distribution $L_{b^{\prime}, b, S S}(z)$ as $b_l$ and $b_u$. \\
    Step 4 & Determine the PCI of $f(\bm{x})$ by:
    \begin{equation}\label{Eq:CIform}
        [ \overline{f}_{\text{DNN}}(\bm{x}) -  b_u/\kappa_{n} ~,~       \overline{f}_{\text{DNN}}(\bm{x}) -  b_l/\kappa_{n} ].
    \end{equation}
    In other words, we take $B_{l} = \overline{f}_{\text{DNN}}(\bm{x}) -  b_u/\kappa_{n}$ and $B_{u} = \overline{f}_{\text{DNN}}(\bm{x}) -  b_l/\kappa_{n}$. 
  \end{tabularx}
\end{algorithm}

Note that to construct the PCI (\ref{Eq:CIform}), the 
values of  $\kappa_n$ and $\kappa_b$ are required.
% According to B1, i.e.,
%$$ \text{Var}(n^\alpha\widehat{f}_{\text{DNN}}(\bm{x})) \rightarrow \sigma^2~ \text{as} ~ n \rightarrow \infty, $$
%by the CLT for triangular arrays, we can find
Recall that $\kappa_n = n^{\frac{1-\beta+2\alpha\beta}{2}}$ and $\kappa_b = n^{\beta\frac{1-\beta+2\alpha\beta}{2}}$.
Although $\beta$ is the practitioner's choice, $\alpha$ is typically
unknown. %\footnote{but can be estimated}  
\cref{Remark:orderkappanb} explains how upper and lower bounds
for $\alpha$ can be used in the PCI construction. 

\begin{Remark}%[The order of $\kappa_n$ and $\kappa_b$]
\label{Remark:orderkappanb}
%Once the sample size $n$ and $\beta$ are determined, $\kappa_n$ and $\kappa_b$ are only affected by the value $\alpha$ which is related to the order of the estimator variance. 
In constructing the PCI (\ref{Eq:CIform}) we can
replace  $\kappa_b$  by a larger value (say $\bar \kappa_b$) and replace $\kappa_n$ by a smaller value (say $\underline{\kappa}_n$)
and still the coverage bound of \cref{Eq:finalCIgoal} would be met. 
%. In other words, we can pretend that the variance decreases faster and slower than its true speed when we fix $\kappa_b$ and $\kappa_n$, respectively. 
From \cref{Theorem: 1}, the fastest rate of the variance decrease is of order $O(n^{-1})$; so $\alpha$ could be as large as $1/2$ in which  $\bar \kappa_b=n^{\frac{\beta}{2}}$. On the other hand, the slowest rate is influenced by $n^{-\frac{\xi}{\xi +d}}$; if we pretend the smoothness of the true model is equal to the input dimension (although it is actually smoother), we can take $\alpha = 1/4 $ to compute $\underline{\kappa}_n= n^{\frac{1-\beta/2}{2}}$. 
\end{Remark}

\section{Predictive inference with the  DNN estimator}\label{Sec:Preinf}
Most of the work in DNN estimation has applications in prediction although
 this is typically point prediction. 
However, as in the estimation case, it is important to be able to 
quantify the accuracy of the point predictors which can be
done via the construction of Prediction Intervals (PI);  see related work of \cite{pan2016bootstrap,wang2021model,zhang2023bootstrap,wu2023bootstrap} on predictive inference with dependent or independent data.

 Consider the problem of predicting a response $Y_{0}$ that is associated
with a regressor value of interest denoted by $\bm{x}_{0}$ and its corresponding prediction interval. %We should notice that the idea of the prediction inference is analogous to taking testing data to measure the performance of one model. 
The $L_2$ optimal point predictor of $Y_{0}$ is $f(\bm{x}_{0})$
which is well approximated by $\bar f_{DNN} (\bm{x}_{0})$
as  \cref{Theorem: SSDNNerror} shows.  
To construct a PI for $Y_{0}$, we need to take the variability of the errors into account since, % rather than relying on the CI since the new response $Y_{0}$ consists of two parts once it is 
 conditionally on $\bm{X}_0= \bm{x}_0$, we have $Y_{0} = f(\bm{x}_0) + \epsilon_0$.

If the model $f$ and the  error 
distribution $F_{\epsilon}$ were both known, we could 
construct a PI which covers $Y_{0}$ with $1-\delta$ confidence level 
as follows:
\begin{equation}\label{Eq:PIform}
    \left[ f(\bm{x}_0) + z_{\epsilon, \delta/2} , f(\bm{x}_0) + z_{\epsilon,1-\delta/2}    \right];
\end{equation}
here $z_{\epsilon,1-\delta/2}$ and $z_{\epsilon, \delta/2}$ are the   $1-\delta/2$ and $\delta/2$ quantile values of $F_{\epsilon}$, respectively. Of course, we do not know the true model $f$ but we may replace it with our scalable subsampling DNN estimator $\overline{f}_{\text{DNN}}$. In addition, $F_{\epsilon}$ is also unknown and must be estimated; a typical estimator is $\widehat{F}_{\epsilon}$ which is the empirical distribution of residuals. 
To elaborate,  we define $\widehat{F}_{\epsilon}$ as follows:

\begin{equation}\label{Eq:fittedres}
\begin{split}
    &\widehat{F}_{\epsilon}(z) := \frac{1}{n} \sum_{i=1}^{n}\mathbbm{1}_{\hat{\epsilon}_{i}\leq z};~\mathbbm{1}_{(\cdot)} \text{is the indicator function.} \\
    & \hat{\epsilon}_{i} = f(\bm{x}_i) - \overline{f}_{\text{DNN}}(\bm{x}_i),~\text{for}~i = 1,\ldots,n. 
\end{split}
\end{equation}

To consistently estimate the error distribution $F_{\epsilon}$, we need to make some mild assumptions on $F_{\epsilon}$, namely:
\begin{itemize}
    \item B2: The error distribution $F_{\epsilon}$ has zero mean and is differentiable on the real line and $\sup_zp_\epsilon(z)<\infty$ were $p_\epsilon(z)$ is the density function of error $\epsilon$.
\end{itemize}

The following Lemma can be proved analogously to the proof of Lemma 4.1 in \cite{wu2023bootstrap}. 

\begin{Lemma}\label{Lemma:5.1}
 Under A1-A5 and B1, we have
%$F_{\epsilon}$ can be estimated by $\widehat{F}_{\epsilon}$ uniformly in the below way:
 $
         \sup_{z}|\widehat{F}_{\epsilon}(z) - F_{\epsilon}(z)|\overset{p}{\to} 0.
$
\end{Lemma}

We can then apply the PI below to approximate
the `oracle' PI of  \cref{Eq:PIform}:
\begin{equation}\label{Eq:estPIform}
     \left[ \overline{f}_{\text{DNN}}(\bm{x}_0) + \hat{z}_{\epsilon,\delta/2} , \overline{f}_{\text{DNN}}(\bm{x}_0) + \hat{z}_{\epsilon,1-\delta/2}    \right];
\end{equation}
here $\hat{z}_{\epsilon,1-\delta/2}$ and $\hat{z}_{\epsilon,\delta/2}$ are the   $1-\delta/2$ and $\delta/2$ quantile values of $\widehat{F}_{\epsilon}$, respectively. To construct this PI in practice, we can rely on \cref{Algo: BootPI} below:

\begin{algorithm}[htbp]
\caption{PI of $Y_0$ conditional on $\bm{x}_0$} 
%\centering
\label{Algo: BootPI}
  %\centering
  \begin{tabularx}{\textwidth}{lX}   
    Step 1 & Train the subagging DNN estimator $\overline{f}_{\text{DNN}}(\cdot)$ and find the empirical distribution of residuals $\widehat{F}_\epsilon$ as \cref{Eq:fittedres}. \\
    Step 2 & Evaluate the subagging DNN estimator at $\bm{x}_0$ to get $\overline{f}_{\text{DNN}}(\bm{x}_0)$.\\
    Step 3 & Determine $\hat{z}_{\epsilon,\delta/2}$ and $\hat{z}_{\epsilon,1-\delta/2}$ by taking lower $\delta/2$ and $1-\delta/2$ quantiles of $\widehat{F}_\epsilon$.   \\
    Step 4 & Construct PI as \cref{Eq:estPIform}.   \\
  \end{tabularx}
\end{algorithm}

We claim that the PI in \cref{Eq:estPIform} is asymptotically valid (conditionally on $\bm{X}_0= \bm{x}_0$), i.e., it satisfies

\begin{equation}\label{Eq:PIstate}
    \mathbb{P}\left( Y_0 \in  \left[ \overline{f}_{\text{DNN}}(\bm{x}_0) + \hat{z}_{\epsilon,\delta/2} , \overline{f}_{\text{DNN}}(\bm{x}_0) + \hat{z}_{\epsilon,1-\delta/2}    \right] \right) \overset{p}{\to} 1 - \delta,
\end{equation}
where the above probability is conditional on $\bm{X}_0= \bm{x}_0$. This statement is guaranteed by \cref{Theorem:PIconver}. To describe it, denote $Y^{*}_0 = \overline{f}_{\text{DNN}}(\bm{x}_0) + \epsilon^*_0$ where $\epsilon^*_0$ has the distribution $\widehat{F}_{\epsilon}$.

\begin{Theorem}\label{Theorem:PIconver}
 Under A1-A5 and B1-B2, the distribution of $Y^{*}_0$
% = \overline{f}_{\text{DNN}}(\bm{x}_0) + \epsilon^*_0$
 converges to the distribution of $Y_0 $
%= f(\bm{x}_0) + \epsilon_0$
 uniformly (in probability), i.e.,   
\begin{equation}
    \sup_{z}\left|F_{Y^*_0|\bm{x}_{0} = \bm{x}_0}(z) - F_{Y_{0}|\bm{x}_{0} = \bm{x}_0}(z)\right|\overset{p}{\to} 0, ~\text{as}~n \to \infty.
\end{equation}
%here $\epsilon^*_0$ has the distribution $\widehat{F}_{\epsilon}$.
\end{Theorem}
Although the PI in \cref{Eq:estPIform} is asymptotically valid, it 
may undercover $Y_{0}$ in the finite sample case. This problem is mainly due to two reasons: (1) PI in \cref{Eq:estPIform} does not take the variability of model estimation into account; and (2) the scale of the error distribution is typically underestimated by the residual distribution with finite samples. For issue (1), we can rely on a so-called pertinent PI which is able to capture the model estimation variability; this pertinence property is crucial, especially for the prediction inference of time series data in which multiple-step ahead forecasting is usually required. For issue (2), we can ``enlarge'' the residual distribution by
basing it on the so-called predictive (as opposed to fitted) residuals. Although the predictive residuals are asymptotically equivalent to the fitted residuals, i.e., $\hat{\epsilon}$ in \cref{Eq:fittedres}, the corresponding PI could have a better coverage rate; see \cite{politis2015model} for the formal definition of pertinent PI and predictive residuals. 

In this paper, due to the computational issues in fitting DNN models, we only build the PI in \cref{Eq:estPIform}. Taking a fairly large enough sample size in \cref{Sec:Sim}, this PI works well, and its empirical coverage rate is only slightly lower than that of the oracle.

\section{Simulations}\label{Sec:Sim}
In this section, we attempt to check the performance of the scalable subagging DNN estimator with simulation examples. More specifically, we consider two aspects of one estimator: (1) Time-complexity, we take the running time of the training stage to measure its complexity for a fixed hyperparameter setting, e.g., fixed number of epochs and batch size; (2) Estimation accuracy, we take empirical MSE (mean square error)/MSPE (mean square prediction error) and empirical coverage rate to measure the accuracy of point estimations/predictions and confidence/prediction intervals. 

\subsection{Simulations on point estimations}\label{Subsec: simpoint}
As shown in \cref{Sec:EstDNN}, the scalable subagging DNN estimator is more computationally efficient but also more accurate meantime compared to the DNN estimator trained with the whole sample size under some mild conditions. Here, we hope to verify such dominating performance with simulated data. To perform simulations, we consider below models:
\begin{itemize}
    \item Model-1: $Y =  \sum_{i=1}^{10}X_i + \epsilon $, where $(X_1,\ldots, X_{10})\sim N(0,\bm{I})$. 
    \item Model-2: $Y = \sum_{i=1}^{10}i\cdot X_i + \epsilon$, where $(X_1,\ldots, X_{10})\sim N(0,\bm{I})$.
    \item Model-3: $Y = X_1^2 + \sin(X_2 + X_3)+ \epsilon$, where $(X_1,X_2,X_3)\sim N(0,\bm{I})$. 
    \item Model-4: $Y = X_1^2 + \sin(X_2 + X_3) + \exp(-|X_4 + X_5|)+ \epsilon$, where $(X_1,X_2,X_3,X_4,X_5)\sim N(0,\bm{I})$;
\end{itemize}
here $\bm{I}$ is an identity matrix with the correct dimension for each model; $\epsilon$ is the standard normal error. We build the DNN estimator with \textit{PyTorch} in \textit{Python}. To train the DNN, we use the stochastic gradient descent algorithm \textit{Adam} developed by \cite{kingma2014adam} with a learning rate $0.01$. In addition, we take the number of epochs and batch size to be 200 and 10 to make the DNN fully trained for the first and iterated subsampling stages, respectively. We use the function \textit{time.time()} in \textit{Python} to compute the running time of the training procedure, namely Training Time.

To be consistent with the folk wisdom, we build $\widehat{f}_{\text{DNN},b,i}$ with a relatively large depth to decrease the bias. Meanwhile, we take the width as large as possible to make its size close to the sample size so that A3 could be satisfied and we are in the under-parameterized region. In order to make a comprehensive comparison between the scalable subsampling DNN (SS-DNN) estimator $\overline{f}_{\text{DNN}}$ and classical DNN estimators, we consider 5 DNN estimators which are trained with the whole sample: 
\begin{itemize}
    \item[(1)] A DNN possesses the same depth and width as $\widehat{f}_{\text{DNN},b,i}$. We denote it ``S-DNN''.
    \item[(2)] A DNN possesses the same depth as $\widehat{f}_{\text{DNN},b,i}$, but a larger width so that its size is close to the sample size. We denote it ``DNN-deep-1''.
    \item[(3)] A DNN possesses the same depth as $\widehat{f}_{\text{DNN},b,i}$, but a larger width so that its size is close to half of the sample size. We denote it ``DNN-deep-2''.
    \item[(4)] A DNN possesses only one hidden layer, but a larger width so that its size is close to the sample size. We denote it ``DNN-wide-1''.
    \item[(5)] A DNN possesses only one hidden layer, but a larger width so that its size is close to half of the sample size. We denote it ``DNN-wide-2''.
\end{itemize}
We deploy DNN (1) to check the performance of a DNN with the same structure as $\widehat{f}_{\text{DNN},b,i}$, but it is trained with the whole dataset. We deploy DNNs (2) - (5) to challenge the scalable subsampling DNN estimator with various wide or deep DNNs. To evaluate the point estimation performance, we apply two empirical MSE criteria:
\begin{equation*}
    \text{MSE-1:}\frac{1}{n} \sum_{i=1}^{n}(\widehat{f}_{\text{DNN}}(\bm{x}_i) - y_i)^2~;~\text{MSE-2:}\frac{1}{n} \sum_{i=1}^{n}(\widehat{f}_{\text{DNN}}(\bm{x}_i) - f(\bm{x}_i))^2;
\end{equation*}
here $\widehat{f}_{\text{DNN}}(\cdot)$ represents different DNN estimators and $f(\cdot)$ is the true regression function; $\{\bm{x}_i, y_i\}_{i=1}^{n}$ are realizations of samples; we call it training data.

An estimator is optimal in MSE-1 criterion if its MSE-1 is closest to the sample variance of errors, namely $\hat{\sigma}^2_{\epsilon} = \frac{1}{n}\sum_{i=1}^{n}\epsilon^2_{i}$; here $\{\epsilon^2_{i}\}_{i=1}^{n}$ are observed error values. An estimator is optimal in the MSE-2 criterion if its MSE-2 is closest to 0. We present MSE-1 and MSE-2 of different estimators in \cref{Table: simresulterror}. In addition, we also present $\hat{\sigma}^2_{\epsilon}$ of the corresponding simulated sample as the benchmark to compare the performance of different estimators according to the MSE-1 criterion. 

Beyond the point estimation measured on training data, we are also interested in the performance of difference DNN estimators on test data. Thus, we generate new samples: $\{\bm{x}_{0,i}, y_{0,i}\}_{i=1}^{N}$; here we take $N = 2\cdot 10^5$ to evaluate the prediction performance. Similarly, we consider two MSPEs and we denote them MSPE-1 and MSPE-2 following:
\begin{equation*}
    \text{MSPE-1:}\frac{1}{N} \sum_{i=1}^{N}(\widehat{f}_{\text{DNN}}(\bm{x}_{0,i}) - y_{0,i})^2~;~\text{MSPE-2:}\frac{1}{N} \sum_{i=1}^{N}(\widehat{f}_{\text{DNN}}(\bm{x}_{0,i}) - f(\bm{x}_{0,i}))^2;
\end{equation*}
we expect that the best estimator on prediction tasks should have the smallest MSPE-2 and the MSPE-1 which is closest to $\hat{\sigma}^2_{\epsilon,0} = \frac{1}{N}\sum_{i=1}^{N}(\epsilon_{0,i})^2$; here $\{\epsilon_{0,i}\}_{i=1}^{N}$ are observed error values for the test data. We present all simulation results in \cref{Table: simresulterror}; here empirical MSE/MSPE and Training Time (in seconds) were computed as averages of 200 replications.
\begin{table}[htbp]
  \centering
  \caption{MSE/MSPE and Training Time (in seconds) of different DNN models on various simulation datasets with error terms}
  \vspace{2pt}
  
    \begin{tabular}{llcccccc}
    \toprule
        \multicolumn{2}{c}{Estimator:} & \thead{ SS-DNN} & \thead{ S-DNN} & \thead{  DNN-deep-1 }& \thead{ DNN-deep-2 } & \thead{ DNN-wide-1  } & \thead{ DNN-wide-2 } \\
    \hline\\
    \multicolumn{4}{l}{Model-1, $n = 10^4$, $\hat{\sigma}^2_{\epsilon}=1.0011$, $\hat{\sigma}^2_{\epsilon,0}=1.0003$  }       &           &       &       &  \\[5pt]

    \multicolumn{2}{l}{Width} & [20,20] & [20,20] & [90,90] & [60,60] & [800] & [400] \\
    \multicolumn{2}{l}{MSE-1} & 1.0034 & 1.0168 &0.9975 & 1.0036 & 1.0136 & 1.0151 \\
    \multicolumn{2}{l}{MSE-2} & 0.1011 & 0.0579 & 0.1039 & 0.0894 & 0.0466 & 0.0433 \\
    \multicolumn{2}{l}{MSPE-1} & 1.1020 & 1.0678 & 1.1299 & 1.1059 & 1.0543 & 1.0487 \\
    \multicolumn{2}{l}{MSPE-2} & 0.1019 & 0.0675 & 0.1296 & 0.1057 & 0.0540 & 0.0484 \\
    \multicolumn{2}{l}{Training Time} & 209   & 225   & 403   & 303   & 373   & 274 \\
          &       &       &       &       &       &       &  \\
        \hline\\
    \multicolumn{4}{l}{Model-2, $n = 10^4$, $\hat{\sigma}^2_{\epsilon}=1.0012$, $\hat{\sigma}^2_{\epsilon,0}=1.0011$  }              &       &       &       &  \\[5pt]
    
    \multicolumn{2}{l}{Width} & [20,20] & [20,20] & [90,90] & [60,60] & [800] & [400] \\
    \multicolumn{2}{l}{MSE-1} & 1.0506 & 1.1355 & 1.1314 & 1.1350 & 1.0768 & 1.0745 \\
     \multicolumn{2}{l}{MSE-2} & 0.1232 & 0.1625 & 0.1889 & 0.1839 & 0.1249 & 0.1194 \\
     \multicolumn{2}{l}{MSPE-1} & 1.1339 & 1.1469 & 1.1841 & 1.1737 & 1.1254 & 1.1237 \\
     \multicolumn{2}{l}{MSPE-2} & 0.1338 & 0.1468 & 0.1841 & 0.1736 & 0.1253 & 0.1238 \\
    \multicolumn{2}{l}{Training Time} & 224   & 240   & 417   & 320   & 376   & 280 \\
          &       &       &       &       &       &       &  \\
    \hline\\
    \multicolumn{4}{l}{Model-3, $n = 10^4$, $\hat{\sigma}^2_{\epsilon}=0.9997$,$\hat{\sigma}^2_{\epsilon,0}=1.0001$}, &                   &       &       &  \\[5pt]
 
    \multicolumn{2}{l}{Width} & [15,15,15] & [15,15,15] & [65,65,65] & [45,45,45] & [2000] & [1000] \\
    \multicolumn{2}{l}{MSE-1} & 1.0014 & 1.0361 & 1.0299 & 1.0308 & 1.0286 & 1.0290 \\
    \multicolumn{2}{l}{MSE-2} & 0.0296 & 0.0536 & 0.0533 & 0.0522 & 0.0426 & 0.0431 \\
    \multicolumn{2}{l}{MSPE-1} &1.0310 & 1.0565 & 1.0572 & 1.0571 & 1.0453 & 1.0449 \\
    \multicolumn{2}{l}{MSPE-2} & 0.0310 & 0.0564 & 0.0572 & 0.0570 & 0.0453 & 0.0449 \\
    \multicolumn{2}{l}{Training Time} & 353   & 379   & 561   & 468   & 483   & 363 \\

           &       &       &       &       &       &       &  \\
    \hline\\
    \multicolumn{4}{l}{Model-4, $n = 10^4$, $\hat{\sigma}^2_{\epsilon}=1.0014$,$\hat{\sigma}^2_{\epsilon,0}=1.0003$}               &       &       &       &  \\[5pt]
 
    \multicolumn{2}{l}{Width} & [15,15,15] & [15,15,15] & [65,65,65] & [45,45,45] & [2000] & [1000] \\
    \multicolumn{2}{l}{MSE-1} & 1.0243 & 1.0488 & 1.0318 & 1.0350 & 1.0457 & 1.0460 \\
    \multicolumn{2}{l}{MSE-2} & 0.0757 & 0.0830 & 0.1076 & 0.0980 & 0.0729 & 0.0728 \\
    \multicolumn{2}{l}{MSPE-1} & 1.0792 & 1.0878 & 1.1117 & 1.1048 & 1.0756 & 1.0752 \\
    \multicolumn{2}{l}{MSPE-2} & 0.0790 & 0.0875 & 0.1114 & 0.1045 & 0.0754 & 0.0749 \\
    \multicolumn{2}{l}{Training Time} & 359   & 376   & 560   & 471   & 551   & 394 \\

               &       &       &       &       &       &       &  \\
    \hline\\
    \multicolumn{4}{l}{Model-4, $n = 2\cdot 10^4$, $\hat{\sigma}^2_{\epsilon}=0.9991$,$\hat{\sigma}^2_{\epsilon}=0.9999$}          &       &       &       &  \\[5pt]
 
    \multicolumn{2}{l}{Width} & [20,20,20] & [20,20,20] & [95,95,95] & [65,65,65] & [2800] & [1400] \\
    \multicolumn{2}{l}{MSE-1} & 1.0093 & 1.0483 & 1.0419 & 1.0438 & 1.0508 & 1.0508 \\
    \multicolumn{2}{l}{MSE-2} & 0.0490 & 0.0653 & 0.0686 & 0.0675 & 0.0635 & 0.0635 \\
    \multicolumn{2}{l}{MSPE-1} & 1.0501 & 1.0669 & 1.0692 & 1.0689 & 1.0622 & 1.0625 \\
    \multicolumn{2}{l}{MSPE-2} & 0.0502 & 0.0670 & 0.0692 & 0.0689 & 0.0623 & 0.0626 \\
    \multicolumn{2}{l}{Training Time} & 748   & 775   & 1684   & 1198   & 1549   & 998 \\

    \bottomrule  
    \end{tabular}%
    \raggedright
    \vspace{2pt}
     \textit{Note:} ``width'' represents the number of neurons of each hidden layer, e.g., [20, 20] means that there are two hidden layers within the DNN and each has 20 number neurons.
     \label{Table: simresulterror}
\end{table}%
We can summarize several notable findings from the simulation results:
\begin{itemize}
    \item $\overline{f}_{\text{DNN}}$ is always the most computationally efficient one, it is even faster than applying a single DNN estimator with the same structure as $\widehat{f}_{\text{DNN},b,i}$ but trained on the whole sample.
    \item According to the MSE-1, $\overline{f}_{\text{DNN}}$ is the most accurate one for all simulations.
    \item According to the MSE-2, $\overline{f}_{\text{DNN}}$ can work best when the data is large enough for Models 3-4 which are non-linear. For Model-2, the performance of $\overline{f}_{\text{DNN}}$ is just slightly worse than the optimal estimator. For Model-1, the performance of $\overline{f}_{\text{DNN}}$ is still worse than the optimal estimator. We guess the reason may be that the Model-1 and Model-2 are linear models. In this case, a wide DNN is sufficient to mimic such a linear relationship.  

    \item For MSPEs, $\overline{f}_{\text{DNN}}$ works slightly worse than the optimal model for Model-1 and Model-2 cases, but it turns out to be the optimal one for Model-3 and Model-4 cases. This phenomenon is consistent with the behavior of MSEs.

    \item The model-selection step for ``wide'' or ``deep'' type DNN estimators is necessary but it is obscure meanwhile; see DNN-wide-2 works better than DNN-wide-1 for the Model-2 MSE case; however, the situation reverses for the Model-3 MSE case. This phenomenon occurs for ``Deep'' type DNN estimators also; see the performance of S-DNN, DNN-deep-1 and DNN-wide-2; there is no single one that beats the others uniformly. For MSPE, we can also find such a reverse phenomenon. On the other hand, by applying the scalable subagging estimator, we can avoid the model-selection difficulty and just make $\widehat{f}_{\text{DNN},b,i}$ deep and large enough. 
\end{itemize}

To analyze the ability of various DNN estimators on estimating regression models solely, we present additional simulation results in \hyperref[Appendix:B]{Appendix: B} where the four models described above do not have error terms, so the MSE-1 and MSE-2 coincide to each other. 

\FloatBarrier

\subsection{Simulations for CI and PI}
We continue using the four models in \cref{Subsec: simpoint} to test the accuracy of multiple confidence and prediction intervals defined in previous sections with scalable subagging DNN estimators. To make sure we have enough subsamples to do iterated subsampling for CI, we take the sample size to be $2\cdot 10^5$, which implies $q = 38$ when $\beta = 0.7$. It further implies that the number of subsamples for the iterated subagging stage is $q = \lfloor n^{\beta(1-\beta)} \rfloor = 12$. For developing the prediction interval, we take the sample size to be $10^4$ or $2\cdot 10^4$. To determine the structure of the subagging DNN estimator, we keep the strategy summarized in the previous subsection, i.e., we make its size as close to the sample size as possible no matter in the first or the iterated subsampling stage. We take the same training setting with \textit{PyTorch} to find $\overline{f}_{\text{DNN}}(\bm{x}_0)$ as we have done in \cref{Subsec: simpoint}.

We call the naive QCI which is determined by the equal-tail quantile of estimations $\{\widehat{f}_{\text{DNN},b,1}(\bm{x}),$ $\ldots,\widehat{f}_{\text{DNN},b,q}(\bm{x})\}$ QCI-1; we should notice that this QCI may be too conservative as we explained in \cref{Example: typeCI}; we call the QCI based on \cref{Eq:CIform} QCI-2; we call the PCI based on \cref{Eq:simplePCI} PCI-1; we call the PCI based on \cref{Eq:twoenlargedPCI} with taking $2\alpha = 0$, PCI-2; we call the PCI based on \cref{Eq:twoenlargedPCI} with taking $2\alpha = 1$, PCI-3; the PI represents the prediction interval defined in \cref{Eq:estPIform}. For all CIs and PI defined in previous sections, they have asymptotically validity conditional on the observation $\bm{X} = \bm{x}$. We attempt to check the conditional coverage rate with simulations for finite sample cases. To achieve this purpose, we fix 10 unchanged test points $\{(y_{t,1},\bm{x}_{t,1}), \ldots,(y_{t,10},\bm{x}_{t,10}) \}$ which are different from training points for each simulation model; these 10 points can be recovered by setting $\textit{numpy.random.seed(0)}$ and generate sample according to the model. 

To evaluate the performance of (conditional) CI for each test point, we repeat the simulation process $K = 500$ times and apply the below formulas to compute the empirical coverage rate (ECR) and empirical length (EL) of different CIs for each test point:

\begin{equation*}
     \text{ECR}_j= \frac{1}{K}\sum_{i=1}^{K}\mathbbm{1}_{f(\bm{x}_{t,j})\in [B_{l,i,j},B_{u,i,j}]}~,~\text{EL}_j= \frac{1}{K}\sum_{i=1}^{K} (B_{u,i,j} -  B_{l,i,j}), \text{for}~j = 1,\ldots, 10;
\end{equation*}
here $f(\bm{x}_{t,j})$ is the true model value evaluated at the $j$-th test data point; $B_{u,i,j}$ and $B_{l,i,j}$ are the corresponding upper and lower bounds of different CIs at the $i$-th replication for the $j$-th test point, respectively. We take two nominal significance levels $\delta=0.05$ and $\delta=0.1$. Simulation results are tabularized in \cref{Table:simresulterrorCI1,Table:simresulterrorCI2}.

To evaluate the performance of (conditional) PI for each test point, the procedure is slightly complicated and we summarize it in below four steps:
\begin{itemize}
       \item[Step 1] Take the sample size $n$ to be $10^4$ or $2\cdot 10^4$; simulate $K = 500$ sample sets: $\{  (y^{(k)}_i,\bm{x}^{(k)}_i)_{i=1}^{n} \}_{k=1}^{K}$ based on one of four simulation models.
    \item[Step 2] For each sample set, train the subsampling DNN estimator and build the prediction interval for 10 test points by:
    $$
    [\overline{f}_{\text{DNN}}(\bm{x}_{t,j}) + \hat{z}_{\delta/2}, \overline{f}_{\text{DNN}}(\bm{x}_{t,j}) + \hat{z}_{1-\delta/2}       ],~\text{for}~ j = 1, \ldots, 10, 
    $$
    where $\hat{z}_{\epsilon,1-\delta/2}$ and $\hat{z}_{\epsilon,\delta/2}$ are the $1-\delta/2$ and $\delta/2$ quantile values of the empirical distribution of the residuals, respectively. 
    \item[Step 3] To check the performance of PIs for test points based on each sample set, simulate $\{y_{s, j} \}_{s = 1}^{M}$ conditional on $\bm{x}_{t,j}$ for $j = 1,\ldots, 10$ pretending the true data-generating model is known and check the empirical coverage rate and empirical length by below formulas:
\begin{equation*}
     \text{ECR}_{i,j}= \frac{1}{M}\sum_{s=1}^{M}\mathbbm{1}_{y_{s,j}\in [B_{l,i,j},B_{u,i,j}]}~,~\text{EL}_{i,j}= B_{u,i,j} -  B_{l,i,j}, \text{for}~j = 1,\ldots, 10; i = 1,\ldots, 500;
\end{equation*}
$B_{l,i,j}$ and $B_{u,i,j}$ are the corresponding upper and lower bounds of PI for the $j$-th test point based on $i$-th sample set defined in Step 2; $M = 3000$.
    
    \item [Step 4] For $j = 1,\ldots,10$, estimate $\mathbb{P}(Y_0\in PI | \bm{X}_0= \bm{x}_{t,j})$ by the average of empirical coverage rate of corresponding (conditional) PI on $K$ sample sets, i.e., $\text{Average}(\text{ECR}_{i,j})$ w.r.t. $i$; estimate length of (conditional) PI for $j$-th test point by $\text{Average}(\text{EL}_{i,j})$ w.r.t. $i$.
\end{itemize}
We still take two nominal significance levels $\delta=0.05$ and $\delta=0.1$. Simulation results are tabularized in \cref{Table: simresulterrorcPI}.
\begin{Remarknn}[Different levels of conditioning]
As explained in the work of \cite{wang2021model}, we have several conditioning levels to measure the performance of PI or CI. What we consider in this paper is $\mathbb{P}_1 :=\mathbb{P}(\cdot |\bm{X}_0= \bm{x}_0)$ which shall be interpreted as the conditional probability on $\bm{X}_0= \bm{x}_0$. If we consider the empirical coverage rate of $\text{ECR}_{i,j}$, it approximates another conditioning level, i.e., $\mathbb{P}_2 := \mathbb{P}(\cdot |\bm{X}_0= \bm{x}_0, (Y_n,\bm{X}_n))$; here $(Y_n,\bm{X}_n)$ represents the whole sample. By Lemma 4 of \cite{wang2021model}, if $A \in \sigma\left(\mathbf{X}_n, \mathbf{Y}_n, X_f, Y_0\right)$ is an arbitrary measurable event, then $\mathbb{E}_{Y_n,\bm{X}_n}\mathbb{P}_2(A) = \mathbb{P}_1(A)$. Besides, $1-\delta$ conditional coverage under $\mathbb{P}_1 $ will imply the marginal coverage $\mathbb{P}_0 :=\mathbb{E}_{\bm{X}}\mathbb{P}_1$. This unconditional coverage is implied by the popular Conformal Prediction method in the machine learning community. Simulation studies show that our CIs and PIs also have great unconditional coverage; see results from \hyperref[Appendix:C]{Appendix: C}.
\end{Remarknn}

We can summarize several findings based on simulation results:
\begin{itemize}
   
    \item For the empirical coverage rate of quantile-type CIs, the naive QCI-1 over-covers true model values as we expect. Also, the corresponding CI length is always larger than the length of QCI-2 and it is actually the largest one among 5 different CIs. On the other hand, the specifically designed QCI-2 returns ECRs that are closer to the specified confidence level than QCI-1. Meanwhile, ECR of QCI-2 is larger than the nominal confidence level for almost all test points since we take $\kappa_n$ and $\kappa_b$ according to the strategy in \cref{Remark:orderkappanb} to enlarge the CI.  

    \item For the empirical coverage rate of pivot-type CIs, although the length of PCI-1 is the shortest one, the ECR of PCI-1 is always less than the nominal confidence level for almost all test points since $C_{\sigma}^2$ may be underestimated and we may have the bias issue in practice. For the PCI-3 whose margin of error is enlarged in a mild way, although its ECR is always larger than PCI-1, it still undercover true model value mostly. For the PCI-2 in which the margin of error is enlarged in a most extreme way, it has a much better performance according to the coverage rate but with a larger CI length as a sacrifice. We claim that the PCI-2 is the optimal CI candidate according to the overall performance based on length and coverage rate. For the QCI-2, we conjecture it will be a good alternative if we have more samples so that $L_{b^{\prime}, b, S S}(x)$ can approximate $J_n(x)$ well in the iterated subsampling stage. 

    \item For the prediction task, all PIs for four models and all test points have almost the same coverage rate and length. All ECRs are slightly less than the nominal confidence level which is not a surprise since we omit the variability in the model estimation and the residual distribution may underestimate the true error distribution for a finite sample case. For the length of PI, all PIs' lengths are close to $2\cdot z_{0.95}$ or $2\cdot z_{0.975}$ since the true error distribution is assumed to be standard normal in simulations and we took equal-tail PI.
\end{itemize}

\begin{table}[htbp]
\centering
  \caption{Empirical Coverage Rate and Empirical Length of different (conditional) CIs with various simulation models}
  \vspace{2pt}
  
    \begin{tabular}{llcccccccccc}
    \toprule
    \multicolumn{2}{c}{Test point: } & 1& 2& 3& 4& 5& 6& 7& 8& 9 & 10  \\
    \hline
    \multicolumn{3}{l}{Model-1, $n = 2\cdot 10^5$  } &   & & & & & &              \\
\multicolumn{3}{l}{ECR} &   & & & & & &              \\[3pt]
 \multicolumn{2}{l}{QCI-1 $\delta = 0.10$} & 1.000 & 1.000 & 1.000 & 1.000 & 1.000 & 1.000 & 1.000 & 1.000 & 1.000 & 1.000 \\ 
 \multicolumn{2}{l}{QCI-1 $\delta = 0.05$} &1.000 & 1.000 & 1.000 & 1.000 & 1.000 & 1.000 & 1.000 & 1.000 & 1.000 & 1.000 \\ 
 \multicolumn{2}{l}{QCI-2 $\delta = 0.10$} & 0.966 & 0.964 & 0.990 & 0.990 & 0.946 & 0.814 & 0.994 & 0.972 & 0.978 & 0.946 \\ 
 \multicolumn{2}{l}{QCI-2 $\delta = 0.05$} &0.988 & 0.984 & 0.998 & 1.000 & 0.984 & 0.932 & 0.998 & 0.984 & 0.994 & 0.976 \\ 
 \multicolumn{2}{l}{PCI-1 $\delta = 0.10$} &0.884 & 0.854 & 0.896 & 0.894 & 0.748 & 0.510 & 0.874 & 0.884 & 0.830 & 0.860 \\ 
 \multicolumn{2}{l}{PCI-1 $\delta = 0.05$} &0.938 & 0.918 & 0.954 & 0.940 & 0.836 & 0.610 & 0.942 & 0.938 & 0.890 & 0.918 \\ 
 \multicolumn{2}{l}{PCI-2 $\delta = 0.10$} &0.892 & 0.986 & 0.996 & 1.000 & 0.784 & 0.954 & 0.998 & 0.908 & 0.876 & 0.998 \\ 
 \multicolumn{2}{l}{PCI-2 $\delta = 0.05$} &0.946 & 0.996 & 0.998 & 1.000 & 0.882 & 0.984 & 1.000 & 0.954 & 0.938 & 1.000 \\ 
 \multicolumn{2}{l}{PCI-3 $\delta = 0.10$} &0.886 & 0.862 & 0.906 & 0.914 & 0.748 & 0.524 & 0.880 & 0.884 & 0.830 & 0.878 \\ 
 \multicolumn{2}{l}{PCI-3 $\delta = 0.05$} &0.938 & 0.922 & 0.958 & 0.960 & 0.836 & 0.616 & 0.948 & 0.938 & 0.890 & 0.928 \\ [5pt]
 
 \multicolumn{3}{l}{EL} &   & & & & & &              \\[3pt]
 \multicolumn{2}{l}{QCI-1 $\delta = 0.10$} & 2.39 & 2.24 & 1.93 & 1.92 & 1.51 & 1.04 & 1.38 & 1.94 & 1.46 & 3.10 \\ 
 \multicolumn{2}{l}{QCI-1 $\delta = 0.05$} & 2.94 & 3.05 & 2.49 & 2.46 & 1.85 & 1.30 & 1.94 & 2.65 & 1.81 & 4.19 \\ 
 \multicolumn{2}{l}{QCI-2 $\delta = 0.10$} & 1.44 & 1.25 & 1.27 & 1.27 & 1.00 & 0.80 & 0.96 & 1.23 & 1.02 & 1.50\\ 
 \multicolumn{2}{l}{QCI-2 $\delta = 0.05$} & 1.73 & 1.50 & 1.54 & 1.54 & 1.20 & 0.97 & 1.16 & 1.48 & 1.24 & 1.81 \\ 
 \multicolumn{2}{l}{PCI-1 $\delta = 0.10$} & 0.39 & 0.37 & 0.32 & 0.31 & 0.24 & 0.17 & 0.23 & 0.32 & 0.24 & 0.51  \\ 
 \multicolumn{2}{l}{PCI-1 $\delta = 0.05$} & 0.46 & 0.45 & 0.38 & 0.37 & 0.29 & 0.20 & 0.28 & 0.38 & 0.28 & 0.61  \\ 
 \multicolumn{2}{l}{PCI-2 $\delta = 0.10$} & 0.40 & 0.53 & 0.57 & 1.07 & 0.25 & 0.31 & 0.51 & 0.34 & 0.26 & 0.96\\ 
 \multicolumn{2}{l}{PCI-2 $\delta = 0.05$} & 0.47 & 0.63 & 0.67 & 1.28 & 0.30 & 0.37 & 0.61 & 0.41 & 0.31 & 1.15  \\ 
 \multicolumn{2}{l}{PCI-3 $\delta = 0.10$} & 0.39 & 0.38 & 0.32 & 0.34 & 0.24 & 0.17 & 0.24 & 0.32 & 0.24 & 0.53  \\ 
 \multicolumn{2}{l}{PCI-3 $\delta = 0.05$} & 0.46 & 0.45 & 0.38 & 0.41 & 0.29 & 0.21 & 0.28 & 0.39 & 0.28 & 0.63 \\ [5pt]

\multicolumn{3}{l}{Model-2, $n = 2\cdot 10^5$  } &   & & & & & &              \\
\multicolumn{3}{l}{ECR} &   & & & & & &              \\[3pt]
\multicolumn{2}{l}{QCI-1 $\delta = 0.10$} & 1.000 & 1.000 & 1.000 & 1.000 & 1.000 & 1.000 & 1.000 & 1.000 & 1.000 & 1.000 \\
\multicolumn{2}{l}{QCI-1 $\delta = 0.05$} & 1.000 & 1.000 & 1.000 & 1.000 & 1.000 & 1.000 & 1.000 & 1.000 & 1.000 & 1.000 \\
\multicolumn{2}{l}{QCI-2 $\delta = 0.10$} &  0.910 & 0.960 & 0.948 & 0.918 & 0.910 & 0.628 & 0.972 & 0.936 & 0.930 & 0.928 \\
\multicolumn{2}{l}{QCI-2 $\delta = 0.05$} & 0.952 & 0.980 & 0.980 & 0.962 & 0.960 & 0.746 & 0.984 & 0.974 & 0.970 & 0.958 \\
\multicolumn{2}{l}{PCI-1 $\delta = 0.10$} & 0.886 & 0.890 & 0.892 & 0.862 & 0.860 & 0.828 & 0.888 & 0.866 & 0.882 & 0.888 \\
\multicolumn{2}{l}{PCI-1 $\delta = 0.05$} & 0.932 & 0.952 & 0.954 & 0.926 & 0.916 & 0.894 & 0.944 & 0.940 & 0.936 & 0.958 \\ 
\multicolumn{2}{l}{PCI-2 $\delta = 0.10$} &  0.904 & 0.974 & 0.990 & 1.000 & 0.880 & 0.958 & 0.982 & 0.896 & 0.902 & 0.998 \\
\multicolumn{2}{l}{PCI-2 $\delta = 0.05$} &  0.946 & 0.984 & 0.996 & 1.000 & 0.940 & 0.990 & 0.984 & 0.960 & 0.952 & 1.000 \\
\multicolumn{2}{l}{PCI-3 $\delta = 0.10$} & 0.886 & 0.892 & 0.900 & 0.892 & 0.860 & 0.830 & 0.896 & 0.868 & 0.882 & 0.920 \\ 
\multicolumn{2}{l}{PCI-3 $\delta = 0.05$} &    0.932 & 0.956 & 0.958 & 0.948 & 0.916 & 0.898 & 0.950 & 0.940 & 0.936 & 0.964 \\

 \multicolumn{3}{l}{EL} &   & & & & & &              \\[3pt]
 \multicolumn{2}{l}{QCI-1 $\delta = 0.10$} & 2.66 & 2.05 & 2.45 & 2.72 & 2.13 & 1.93 & 2.05 & 2.48 & 2.55 & 3.17 \\ 
 \multicolumn{2}{l}{QCI-1 $\delta = 0.05$} & 3.31 & 2.60 & 3.04 & 3.35 & 2.58 & 2.35 & 2.75 & 3.04 & 3.09 & 4.21 \\ 
 \multicolumn{2}{l}{QCI-2 $\delta = 0.10$} & 1.19 & 1.08 & 1.15 & 1.26 & 1.04 & 0.92 & 1.15 & 1.18 & 1.19 & 1.33 \\  
 \multicolumn{2}{l}{QCI-2 $\delta = 0.05$} & 1.44 & 1.30 & 1.39 & 1.52 & 1.25 & 1.10 & 1.39 & 1.43 & 1.44 & 1.60 \\ 
 \multicolumn{2}{l}{PCI-1 $\delta = 0.10$} & 0.43 & 0.33 & 0.39 & 0.44 & 0.34 & 0.31 & 0.34 & 0.40 & 0.41 & 0.52 \\ 
 \multicolumn{2}{l}{PCI-1 $\delta = 0.05$} & 0.51 & 0.40 & 0.47 & 0.52 & 0.41 & 0.37 & 0.40 & 0.48 & 0.49 & 0.61 \\
 \multicolumn{2}{l}{PCI-2 $\delta = 0.10$} &  0.44 & 0.48 & 0.61 & 1.12 & 0.35 & 0.39 & 0.56 & 0.42 & 0.42 & 0.96 \\ 
 \multicolumn{2}{l}{PCI-2 $\delta = 0.05$} &  0.52 & 0.57 & 0.73 & 1.33 & 0.41 & 0.47 & 0.67 & 0.50 & 0.50 & 1.14 \\
 \multicolumn{2}{l}{PCI-3 $\delta = 0.10$} & 0.43 & 0.33 & 0.40 & 0.47 & 0.34 & 0.31 & 0.34 & 0.40 & 0.41 & 0.53 \\
 \multicolumn{2}{l}{PCI-3 $\delta = 0.05$} &   0.51 & 0.40 & 0.48 & 0.55 & 0.41 & 0.37 & 0.41 & 0.48 & 0.49 & 0.63 \\ [5pt]
    \bottomrule  
    \end{tabular}%

     \label{Table:simresulterrorCI1}
\end{table}%

\clearpage

\begin{table}[htbp]
\centering
  \caption{Empirical Coverage Rate and Empirical Length of different (conditional) CIs with various simulation models}
  \vspace{2pt}
  
    \begin{tabular}{llcccccccccc}
    \toprule
    \multicolumn{2}{c}{Test point: } & 1& 2& 3& 4& 5& 6& 7& 8& 9 & 10  \\
    \hline
    \multicolumn{3}{l}{Model-3, $n = 2\cdot 10^5$  } &   & & & & & &              \\
\multicolumn{3}{l}{ECR} &   & & & & & &              \\[3pt]
 \multicolumn{2}{l}{QCI-1 $\delta = 0.10$} & 1.000 & 1.000 & 1.000 & 1.000 & 1.000 & 1.000 & 1.000 & 1.000 & 1.000 & 1.000 \\ 
 \multicolumn{2}{l}{QCI-1 $\delta = 0.05$} & 1.000 & 1.000 & 1.000 & 1.000 & 1.000 & 1.000 & 1.000 & 1.000 & 1.000 & 1.000 \\
 \multicolumn{2}{l}{QCI-2 $\delta = 0.10$} & 0.988 & 0.996 & 1.000 & 0.996 & 0.976 & 0.962 & 0.990 & 0.992 & 0.998 & 0.994 \\ 
 \multicolumn{2}{l}{QCI-2 $\delta = 0.05$} & 0.998 & 1.000 & 1.000 & 0.998 & 0.996 & 0.974 & 0.996 & 0.998 & 1.000 & 1.000 \\
 \multicolumn{2}{l}{PCI-1 $\delta = 0.10$} & 0.788 & 0.868 & 0.638 & 0.862 & 0.704 & 0.810 & 0.868 & 0.846 & 0.838 & 0.842 \\ 
 \multicolumn{2}{l}{PCI-1 $\delta = 0.05$} & 0.858 & 0.920 & 0.728 & 0.920 & 0.804 & 0.880 & 0.930 & 0.918 & 0.914 & 0.920 \\
 \multicolumn{2}{l}{PCI-2 $\delta = 0.10$} & 1.000 & 0.884 & 1.000 & 1.000 & 1.000 & 0.902 & 0.908 & 0.978 & 0.876 & 0.970 \\
 \multicolumn{2}{l}{PCI-2 $\delta = 0.05$} & 1.000 & 0.938 & 1.000 & 1.000 & 1.000 & 0.948 & 0.964 & 0.998 & 0.932 & 0.988 \\ 
 \multicolumn{2}{l}{PCI-3 $\delta = 0.10$} &  0.860 & 0.870 & 0.754 & 0.870 & 0.726 & 0.810 & 0.868 & 0.852 & 0.840 & 0.844 \\
 \multicolumn{2}{l}{PCI-3 $\delta = 0.05$} &  0.920 & 0.920 & 0.842 & 0.924 & 0.820 & 0.882 & 0.930 & 0.922 & 0.914 & 0.922 \\[5pt]

 \multicolumn{3}{l}{EL} &   & & & & & &              \\[3pt]
 \multicolumn{2}{l}{QCI-1 $\delta = 0.10$} &  1.24 & 1.22 & 0.41 & 0.47 & 0.77 & 2.39 & 1.64 & 1.10 & 0.91 & 0.78 \\
 \multicolumn{2}{l}{QCI-1 $\delta = 0.05$} &  1.51 & 1.47 & 0.51 & 0.57 & 0.93 & 2.94 & 2.00 & 1.32 & 1.12 & 0.94 \\
 \multicolumn{2}{l}{QCI-2 $\delta = 0.10$} &  0.85 & 0.97 & 0.39 & 0.43 & 0.62 & 1.52 & 1.31 & 0.80 & 0.74 & 0.62 \\
 \multicolumn{2}{l}{QCI-2 $\delta = 0.05$} &  1.03 & 1.16 & 0.47 & 0.52 & 0.75 & 1.83 & 1.57 & 0.97 & 0.90 & 0.76 \\
 \multicolumn{2}{l}{PCI-1 $\delta = 0.10$} &  0.20 & 0.19 & 0.07 & 0.07 & 0.12 & 0.39 & 0.26 & 0.18 & 0.15 & 0.12 \\ 
 \multicolumn{2}{l}{PCI-1 $\delta = 0.05$} &  0.24 & 0.23 & 0.08 & 0.09 & 0.15 & 0.46 & 0.31 & 0.21 & 0.17 & 0.15 \\ 
 \multicolumn{2}{l}{PCI-2 $\delta = 0.10$} &  1.22 & 0.21 & 0.76 & 0.20 & 0.45 & 0.53 & 0.29 & 0.27 & 0.16 & 0.20 \\ 
 \multicolumn{2}{l}{PCI-2 $\delta = 0.05$} &  1.46 & 0.25 & 0.90 & 0.23 & 0.53 & 0.63 & 0.35 & 0.33 & 0.20 & 0.24 \\
 \multicolumn{2}{l}{PCI-3 $\delta = 0.10$} &  0.24 & 0.20 & 0.08 & 0.08 & 0.13 & 0.39 & 0.26 & 0.18 & 0.15 & 0.13 \\
 \multicolumn{2}{l}{PCI-3 $\delta = 0.05$} &  0.28 & 0.23 & 0.10 & 0.09 & 0.15 & 0.47 & 0.31 & 0.21 & 0.17 & 0.15 \\ [5pt]

 \multicolumn{3}{l}{Model-4, $n = 2\cdot 10^5$  } &   & & & & & &              \\
 \multicolumn{3}{l}{ECR} &   & & & & & &              \\[3pt]
 \multicolumn{2}{l}{QCI-1 $\delta = 0.10$} &  1.000 & 1.000 & 1.000 & 1.000 & 1.000 & 1.000 & 1.000 & 1.000 & 1.000 & 1.000 \\
 \multicolumn{2}{l}{QCI-1 $\delta = 0.05$} & 1.000 & 1.000 & 1.000 & 1.000 & 1.000 & 1.000 & 1.000 & 1.000 & 1.000 & 1.000 \\
 \multicolumn{2}{l}{QCI-2 $\delta = 0.10$} &0.738 & 1.000 & 0.996 & 0.910 & 0.994 & 0.924 & 0.982 & 0.998 & 0.988 & 0.864 \\
 \multicolumn{2}{l}{QCI-2 $\delta = 0.05$} &0.852 & 1.000 & 0.998 & 0.966 & 0.998 & 0.968 & 0.994 & 1.000 & 0.996 & 0.940 \\
 \multicolumn{2}{l}{PCI-1 $\delta = 0.10$} &0.878 & 0.832 & 0.870 & 0.856 & 0.870 & 0.868 & 0.860 & 0.872 & 0.494 & 0.662 \\
 \multicolumn{2}{l}{PCI-1 $\delta = 0.05$} &0.932 & 0.902 & 0.940 & 0.912 & 0.928 & 0.938 & 0.910 & 0.932 & 0.590 & 0.776 \\
 \multicolumn{2}{l}{PCI-2 $\delta = 0.10$} &0.962 & 1.000 & 0.948 & 0.984 & 0.998 & 0.894 & 1.000 & 0.998 & 0.998 & 0.998 \\
 \multicolumn{2}{l}{PCI-2 $\delta = 0.05$} &0.980 & 1.000 & 0.986 & 0.996 & 0.998 & 0.948 & 1.000 & 0.998 & 1.000 & 1.000 \\
 \multicolumn{2}{l}{PCI-3 $\delta = 0.10$} & 0.878 & 0.898 & 0.870 & 0.858 & 0.884 & 0.868 & 0.888 & 0.876 & 0.522 & 0.664 \\
 \multicolumn{2}{l}{PCI-3 $\delta = 0.05$} & 0.936 & 0.948 & 0.942 & 0.920 & 0.938 & 0.938 & 0.936 & 0.940 & 0.614 & 0.780 \\ [5pt]

 \multicolumn{3}{l}{EL} &   & & & & & &              \\[3pt]
\multicolumn{2}{l}{QCI-1 $\delta = 0.10$} &2.90 & 0.63 & 0.79 & 1.50 & 2.35 & 1.16 & 1.09 & 1.99 & 1.21 & 0.69 \\
 \multicolumn{2}{l}{QCI-1 $\delta = 0.05$} &3.71 & 0.79 & 1.00 & 1.91 & 2.92 & 1.48 & 1.35 & 2.48 & 1.49 & 0.89 \\
 \multicolumn{2}{l}{QCI-2 $\delta = 0.10$} &1.74 & 0.61 & 0.70 & 1.14 & 1.67 & 0.80 & 0.85 & 1.38 & 0.89 & 0.58 \\
 \multicolumn{2}{l}{QCI-2 $\delta = 0.05$} &2.10 & 0.74 & 0.84 & 1.38 & 2.02 & 0.97 & 1.03 & 1.68 & 1.08 & 0.70 \\
 \multicolumn{2}{l}{PCI-1 $\delta = 0.10$} &0.47 & 0.10 & 0.13 & 0.24 & 0.38 & 0.19 & 0.18 & 0.32 & 0.20 & 0.11 \\
 \multicolumn{2}{l}{PCI-1 $\delta = 0.05$} &0.56 & 0.12 & 0.15 & 0.29 & 0.45 & 0.22 & 0.21 & 0.39 & 0.23 & 0.13 \\
 \multicolumn{2}{l}{PCI-2 $\delta = 0.10$} &0.69 & 0.78 & 0.17 & 0.45 & 0.85 & 0.20 & 0.76 & 0.54 & 0.60 & 0.24 \\
 \multicolumn{2}{l}{PCI-2 $\delta = 0.05$} &0.82 & 0.93 & 0.20 & 0.54 & 1.01 & 0.24 & 0.90 & 0.64 & 0.72 & 0.29 \\
 \multicolumn{2}{l}{PCI-3 $\delta = 0.10$} &0.48 & 0.12 & 0.13 & 0.24 & 0.39 & 0.19 & 0.19 & 0.33 & 0.20 & 0.11 \\
 \multicolumn{2}{l}{PCI-3 $\delta = 0.05$} &0.57 & 0.14 & 0.15 & 0.29 & 0.47 & 0.22 & 0.23 & 0.39 & 0.24 & 0.14 \\

    \bottomrule  
    \end{tabular}%

     \label{Table:simresulterrorCI2}
\end{table}%

\begin{table}[htbp]
\centering
  \caption{Empirical Coverage Rate and Empirical Length of (conditional) PIs  with various simulation models}
  \vspace{2pt}
  
    \begin{tabular}{llcccccccccc}
    \toprule
    \multicolumn{2}{c}{Test point: } & 1& 2& 3& 4& 5& 6& 7& 8& 9 & 10  \\
    \hline
    \multicolumn{3}{l}{$n  = 10^4$ } &   & & & & & &              \\[5pt]
    \multicolumn{3}{l}{Model-1: } &  &  & & &   & & & &             \\
    \multicolumn{5}{l}{EL = 3.28, $\delta = 0.10$; EL = 3.91, $\delta = 0.05$  } &  &  & & &   & &            \\
     \multicolumn{2}{l}{ECR $\delta = 0.10$:} & 0.870 & 0.877 & 0.877 & 0.873 & 0.884 & 0.891 & 0.886 & 0.878 & 0.886 & 0.865 \\ 
     \multicolumn{2}{l}{ECR $\delta = 0.05$:} & 0.929 & 0.934 & 0.934 & 0.931 & 0.939 & 0.944 & 0.940 & 0.935 & 0.940 & 0.925 \\ [5pt]

     \multicolumn{3}{l}{Model-2:   } &  &  & & &               \\
     \multicolumn{5}{l}{EL = 3.35, $\delta = 0.10$; EL = 4.00, $\delta = 0.05$  } &  &  & & &   & &            \\
     \multicolumn{2}{l}{ECR $\delta = 0.10$:} & 0.880 & 0.882 & 0.883 & 0.879 & 0.889 & 0.892 & 0.888 & 0.881 & 0.885 & 0.869 \\ 
     \multicolumn{2}{l}{ECR $\delta = 0.05$:} & 0.936 & 0.938 & 0.938 & 0.935 & 0.942 & 0.945 & 0.942 & 0.937 & 0.939 & 0.928 \\ [5pt]

     \multicolumn{7}{l}{Model-3:  } &  &  & & &               \\
     \multicolumn{5}{l}{EL = 3.29, $\delta = 0.10$; EL = 3.93, $\delta = 0.05$  } &  &  & & &   & &            \\
     \multicolumn{2}{l}{ECR $\delta = 0.10$:} & 0.896 & 0.894 & 0.898 & 0.899 & 0.897 & 0.880 & 0.889 & 0.895 & 0.896 & 0.897 \\ 
     \multicolumn{2}{l}{ECR $\delta = 0.05$:} & 0.948 & 0.946 & 0.949 & 0.949 & 0.948 & 0.936 & 0.943 & 0.947 & 0.947 & 0.949 \\ [5pt]

     \multicolumn{7}{l}{Model-4:  } &  &  & & &               \\
     \multicolumn{5}{l}{EL = 3.32, $\delta = 0.10$; EL = 3.96, $\delta = 0.05$  } &  &  & & &   & &            \\
     \multicolumn{2}{l}{ECR $\delta = 0.10$:} & 0.830 & 0.900 & 0.900 & 0.889 & 0.882 & 0.898 & 0.897 & 0.889 & 0.895 & 0.899  \\ 
     \multicolumn{2}{l}{ECR $\delta = 0.05$:} & 0.901 & 0.951 & 0.950 & 0.943 & 0.938 & 0.949 & 0.949 & 0.943 & 0.947 & 0.950\\ [5pt]

     \multicolumn{3}{l}{$n  = 2\cdot 10^4$ } &   & & & & & & \\ [5pt]
         \multicolumn{3}{l}{Model-1: } &  &  & & &   & & & &             \\
    \multicolumn{5}{l}{EL = 3.28, $\delta = 0.10$; EL = 3.91, $\delta = 0.05$  } &  &  & & &   & &            \\
     \multicolumn{2}{l}{ECR $\delta = 0.10$:} & 0.885 & 0.886 & 0.887 & 0.886 & 0.892 & 0.895 & 0.892 & 0.887 & 0.891 & 0.880 \\ 
     \multicolumn{2}{l}{ECR $\delta = 0.05$:} & 0.939 & 0.941 & 0.941 & 0.940 & 0.945 & 0.946 & 0.945 & 0.941 & 0.943 & 0.936 \\ [5pt]

      \multicolumn{3}{l}{Model-2:   } &  &  & & &               \\
     \multicolumn{5}{l}{EL = 3.32, $\delta = 0.10$; EL = 3.95, $\delta = 0.05$  } &  &  & & &   & &            \\
     \multicolumn{2}{l}{ECR $\delta = 0.10$:} & 0.887 & 0.891 & 0.889 & 0.889 & 0.893 & 0.895 & 0.894 & 0.889 & 0.891 & 0.882 \\ 
     \multicolumn{2}{l}{ECR $\delta = 0.05$:} & 0.941 & 0.944 & 0.942 & 0.943 & 0.945 & 0.947 & 0.946 & 0.943 & 0.944 & 0.938 \\ [5pt]

      \multicolumn{7}{l}{Model-3:  } &  &  & & &               \\
     \multicolumn{5}{l}{EL = 3.29, $\delta = 0.10$; EL = 3.92, $\delta = 0.05$  } &  &  & & &   & &            \\
     \multicolumn{2}{l}{ECR $\delta = 0.10$:} & 0.897 & 0.896 & 0.899 & 0.899 & 0.899 & 0.889 & 0.893 & 0.897 & 0.897 & 0.899 \\ 
     \multicolumn{2}{l}{ECR $\delta = 0.05$:} & 0.948 & 0.947 & 0.950 & 0.949 & 0.949 & 0.943 & 0.945 & 0.948 & 0.948 & 0.949 \\ [5pt]

          \multicolumn{7}{l}{Model-4:  } &  &  & & &               \\
     \multicolumn{5}{l}{EL = 3.30, $\delta = 0.10$; EL = 3.94, $\delta = 0.05$  } &  &  & & &   & &            \\
     \multicolumn{2}{l}{ECR $\delta = 0.10$:} & 0.858 & 0.900 & 0.899 & 0.892 & 0.887 & 0.897 & 0.898 & 0.891 & 0.895 & 0.899  \\ 
     \multicolumn{2}{l}{ECR $\delta = 0.05$:} & 0.921 & 0.950 & 0.950 & 0.945 & 0.941 & 0.948 & 0.949 & 0.944 & 0.947 & 0.949 \\ [5pt]

    \bottomrule  
    \end{tabular}%
     \label{Table: simresulterrorcPI}
\end{table}%

\FloatBarrier

\section{Conclusions}\label{Sec:Conclu}
In this paper, we revisit the error bound of fully connected DNN with the ReLU activation function on estimating regression models. By taking into account the latest DNN approximation results, we improve the current error bound. Under some mild conditions, we show that the error bound of the DNN estimator may be further improved by applying the scalable subsampling technique. As a result, the scalable subsampling DNN estimator is computationally efficient without sacrificing accuracy. The theoretical result is verified by extensive simulation results with various linear or non-linear regression models. 

Beyond the error analysis for point estimations and point predictions, we propose different approaches to build asymptotically valid confidence and prediction intervals. More specifically, to overcome the undercoverage issue of CIs with finite samples, we consider several methods to enlarge the CI. As shown by simulations, our point estimations/predictions and confidence/prediction intervals based on scalable subsampling work well in practice. All in all, the scalable subsampling DNN estimator offers the complete package in terms of statistical inference, i.e., 
(a) computational efficiency;
(b) point estimation/prediction accuracy; and
(c) allowing for the construction of practically useful 
confidence and prediction intervals.
% The additional benefit is that the structure selection stage of building DNN estimator can be alleviated. 

\newpage

\appendix
\section*{\textsc{Appendix A: Proofs}}\label{Appendix:A}
\begin{proof}[\textbf{\textsc{Proof of Theorem 4.1}}]
This result can be easily shown based on the proof of Theorem 1 in the work of \cite{farrell2021deep}. We take the intermediate result from the final step of their proof: With probability at least $1-\exp(-\gamma)$, 
\begin{equation}\label{Eq:proof1bound}
\left\|\widehat{f}_{\text{DNN}}-f\right\|_{L_2(X)} \leq C \left(\sqrt{\frac{H^2 L^2 \log \left(H^2 L\right)}{n} \log n}+\sqrt{\frac{\log \log n+\gamma}{n}}+\epsilon_n\right),
\end{equation}
where $C$ is an appropriate constant; in this proof, $C$ represents appropriate constants and its meaning may change according to the context; $\epsilon_n = \left\|f_{\text{DNN}}-f\right\|_{\infty}$; $f_{\text{DNN}} =  \arg\min_{f_{\theta}\in\mathcal{F}_{\text{DNN}}}\left\|f_{\theta}-f\right\|_{\infty}$. By Theorem 3.1 of \cite{yarotsky2020phase} and Lemma 1 of \cite{farrell2021deep}, we can conclude that there is a standard fully connected DNN whose depth and width satisfy below inequalities:
\begin{equation}\label{Eq:HLbound}
\begin{split}
  & H \leq C \epsilon_n^{-\frac{d}{\xi}}\log \left(1 / \epsilon_n\right), 
\\ & L \leq C \cdot\log \left(1 / \epsilon_n\right),
\end{split}
\end{equation}
for any $\epsilon_n$; Furthermore, we can find the upper bound of $H^2 L^2 \log \left(H^2 L\right)$ based on \cref{Eq:HLbound}:
\begin{equation*}
    H^2 L^2 \log \left(H^2 L\right) \leq C \cdot \epsilon_n^{-\frac{2 d}{\xi}}\left(\log \left(1 / \epsilon_n\right)\right)^5.
\end{equation*}
Subsequently, we rewrite the \cref{Eq:proof1bound} as below:
\begin{equation}
    \left\|\widehat{f}_{\text{DNN}}-f\right\|_{L_2(X)} \leq C\left(\sqrt{\frac{\epsilon_n^{-\frac{2 d}{\xi}}\left(\log \left(1 / \epsilon_n\right)\right)^5}{n} \log n}+\sqrt{\frac{\log \log n+\gamma}{n}}+\epsilon_n\right).
\end{equation}
To optimize the bound, we can choose $\epsilon_n=n^{-\frac{\xi}{2(\xi+d)}}, H = \Theta( n^{\frac{d}{2(\xi+d)}} \log n ), L = \Theta( \log n) $. This gives:
\begin{equation}
      \left\|\widehat{f}_{\text{DNN}}-f\right\|_{L_2(X)}  \leq C\left(n^{-\frac{\xi}{2(\xi+d)}} \log ^3 n+\sqrt{\frac{\log \log n+\gamma}{n}}\right).
\end{equation}
As a result, we get:
\begin{equation}
    \left\|\widehat{f}_{\text{DNN}}-f\right\|_{L_2(X)}^2  \leq C\left(n^{-\frac{\xi}{(\xi+d)}} \log ^6 n+ \frac{\log \log n+\gamma}{n}\right).
\end{equation}
Finally, we take $\gamma = n^{\frac{d}{d+\xi}}\log^6(n)$, which implies \cref{Theorem: 1}. 

\end{proof}

\begin{proof}[\textbf{\textsc{Proof of Theorem 4.2}}]
Under A1-A5, we can analyze the expected square error for the subagging DNN estimator as below:
\begin{equation}\label{Eq:maindecomp}
\begin{split}
    &\mathbb{E}(\overline{f}_{\text{DNN}}(\bm{X})  - f(\bm{X}))^2\\
    & = \mathbb{E}\left[ \frac{1}{q}\sum_{i=1}^q \widehat{f}_{\text{DNN},b,i}(\bm{X}) - f(\bm{X})  \right]^2 \\
    & = \frac{1}{q^2}\mathbb{E}\left[ \sum_{i=1}^q\left( \widehat{f}_{\text{DNN},b,i}(\bm{X}) - f(\bm{X})\right)  \right]^2 \\
    & = \frac{1}{q^2}\mathbb{E} \left[ \sum_{i=1}^{q} \left( \widehat{f}_{\text{DNN},b,i}(\bm{X}) - f(\bm{X})\right)^2  \right ] + \frac{1}{q^2}\mathbb{E}\left[ \sum_{i,j,i\neq j} \left( \widehat{f}_{\text{DNN},b,i}(\bm{X}) - f(\bm{X})\right) \cdot \left( \widehat{f}_{\text{DNN},b,j}(\bm{X}) - f(\bm{X})\right)  \right].
\end{split}
\end{equation}
For the first term on the r.h.s. of \cref{Eq:maindecomp}, by the error bound ignoring the slowly varying term, we can get:
\begin{equation}\label{Eq:firstterm}
\begin{split}
    \frac{1}{q^2}\mathbb{E} \left[ \sum_{i=1}^{q} \left( \widehat{f}_{\text{DNN},b,i}(\bm{X}) - f(\bm{X})\right)^2  \right ] &\leq \frac{1}{q^2} \cdot q \cdot O\left(n^{-\frac{\beta\xi}{\xi + d}}\right) \\
    & = \frac{1}{q} O\left(\frac{1}{n^{\frac{\beta\xi}{\xi + d}}}\right)\\
    & = O\left(\frac{1}{n^{\frac{\beta\xi}{\xi + d} + 1 - \beta}}\right);
\end{split}
\end{equation}
this is satisfied with at least probability $(1 - \exp(-n^{\frac{d}{\xi+d}}\log^6n))^q$.

Ideally, we hope $\beta$ can take a small value to improve the error bound for \cref{Eq:firstterm}. However, it is restricted to do this since the bias of the subagging estimator will get increased once we take $\beta$ smaller and smaller. Thus, we need to consider the second term on the r.h.s. of \cref{Eq:maindecomp}. Start by considering on specific pair:
\begin{equation}\label{Eq:secondterm}
    \begin{split}
        &\mathbb{E}\left[\left( \widehat{f}_{\text{DNN},b,i}(\bm{X}) - f(\bm{X})\right) \cdot \left( \widehat{f}_{\text{DNN},b,j}(\bm{X}) - f(\bm{X})\right)\right] \\
        & = \mathbb{E} \left[ \mathbb{E}\left[\left( \widehat{f}_{\text{DNN},b,i}(\bm{X}) - f(\bm{X})\right) \cdot \left( \widehat{f}_{\text{DNN},b,j}(\bm{X}) - f(\bm{X})\right) \bigg| \bm{X} \right]  \right] \\
        & = \mathbb{E} \left[ \mathbb{E}\left[\left( \widehat{f}_{\text{DNN},b,i}(\bm{X}) - f(\bm{X})\right) 
 \bigg| \bm{X} \right] \cdot \mathbb{E}\left[ \left( \widehat{f}_{\text{DNN},b,j}(\bm{X}) - f(\bm{X})\right) \bigg| \bm{X} \right]  \right].
    \end{split}
\end{equation}
The last equality is due to the independence between subsample $B_i$ and $B_j$. As we mentioned in the main text, we face difficulty in determining the rate of the bias of the subagging estimator. Thus, A4 and A5 are used to make additional assumptions on the bias term. We present A4 as below:
\begin{equation*}
    \mathbb{E}(\widehat{f}_{\text{DNN}}(\bm{x})  - f(\bm{x})) = O(n^{- \Lambda/2})~;~\mathbb{E}(\widehat{f}_{\text{DNN},b,i}(\bm{x})  - f(\bm{x})) = O(n^{- \beta\Lambda/2}).
\end{equation*}
A5 then requires the bias order of $\widehat{f}_{\text{DNN}}$ satisfies the inequality: $\Lambda > \frac{\xi}{\xi + d}$.   

Then, we can find the order of \cref{Eq:secondterm} is:
\begin{equation*}
    \mathbb{E}\left[\left( \widehat{f}_{\text{DNN},b,i}(\bm{X}) - f(\bm{X})\right) \cdot \left( \widehat{f}_{\text{DNN},b,j}(\bm{X}) - f(\bm{X})\right)\right] = O(n^{-\beta\Lambda}). 
\end{equation*}

Combine these two pieces, we can analyze \cref{Eq:maindecomp}:
\begin{equation}
    \begin{split}
        &\mathbb{E}(\overline{f}_{\text{DNN}}(\bm{X})  - f(\bm{X}))^2\\
        & \leq O\left(\frac{1}{n^{\frac{\beta\xi}{\xi + d} + 1 - \beta}}\right) + 2 \cdot \frac{1}{q^2}\cdot {{q}\choose{2}} \cdot O\left(\frac{1}{n^{\beta\Lambda}}\right)\\
        & = O\left(\frac{1}{n^{\frac{\beta\xi}{\xi + d} + 1 - \beta}}\right) +  O\left(\frac{1}{n^{\beta\Lambda}}\right).
    \end{split}
\end{equation}

If the bias term is more negligible than the other term, i.e., 
\begin{equation*}
    \beta\Lambda \geq  \frac{\beta\xi}{\xi + d} + 1 - \beta,~i.e., \beta \geq \frac{1}{1+ \Lambda - \frac{\xi}{\xi + d}}.
\end{equation*}

The above lower bound satisfies the requirement of $\beta$ being positive. %Also, we can find:
% \begin{equation*}
%     \frac{\beta\xi}{\xi + d} + 1 - \beta \geq \frac{\xi}{\xi + d}, ~\text{if}~ \beta \leq 1.
% \end{equation*}
Then, $\Lambda$ needs to be larger than $\frac{\xi}{\xi +d}$ to make sure the lower bound of $\beta$ is less than 1 which is satisfied due to A5. Meanwhile, we want to take $\beta$ as small as possible, i.e., $\beta = \frac{1}{1+ \Lambda - \frac{\xi}{\xi + d}}$. This results in the error bound below:
\begin{equation*}
    \mathbb{E}(\overline{f}_{\text{DNN}}(\bm{X})  - f(\bm{X}))^2 \leq O\left(n^{\frac{-\Lambda}{\Lambda + \frac{d}{\xi + d}}} \right).
\end{equation*}

The fact that $\frac{\Lambda}{\Lambda + \frac{d}{\xi + d}}$ is larger than $\frac{\xi}{\xi +d}$ is guaranteed by the requirement that $\Lambda > \frac{\xi}{\xi +d}$, i.e., A5 again.

\end{proof}

\begin{proof}[\textbf{\textsc{Proof of Theorem 5.1}}]

Since error $\epsilon_0$ and $\epsilon^*_0$ are independent to $\bm{x}_0$, we actually have $ \sup_{z}\left| F_{\epsilon^{*}_0|\bm{X}_0= \bm{x}_0}(z) -  F_{\epsilon_0|\bm{X}_0= \bm{x}_0}(z) \right| \overset{p}{\to}0$ based on \cref{Lemma:5.1}. Thus, we can write:
\begin{equation}
    \sup_{z} | \mathbb{P}(Y^*_0 - \overline{f}_{\text{DNN}}(\bm{x}_0)\leq z) - \mathbb{P}( Y_{0}- f(\bm{x}_0) \leq z)  | \overset{p}{\to} 0 ,
\end{equation}
where $\mathbb{P}(\cdot)$ represents $\mathbb{P}(\cdot | \bm{X}_0= \bm{x}_0)$. We can start by considering the below expression:
\begin{equation}
\begin{split}
    &\sup_{z} | \mathbb{P}(Y^*_0 - f(\bm{x}_0)\leq z) - \mathbb{P}(  Y_{0}- f(\bm{x}_0) \leq z)  | \\
    &= \sup_{z} | \mathbb{P}(Y^*_0 - f(\bm{x}_0)\leq z) - \mathbb{P}(Y_0^* - \overline{f}_{\text{DNN}}(\bm{x}_0)\leq z)+ \mathbb{P}(Y_0^* - \overline{f}_{\text{DNN}}(\bm{x}_0)\leq z)- \mathbb{P}(  Y_{0}- f(\bm{x}_0) \leq z)  | \\
    &\leq \sup_{z} | \mathbb{P}(Y^*_0 - f(\bm{x}_0)\leq z) - \mathbb{P}(Y_0^* - \overline{f}_{\text{DNN}}(\bm{x}_0)\leq z) | + \sup_{z}|\mathbb{P}(Y_0^* - \overline{f}_{\text{DNN}}(\bm{x}_0)\leq z)- \mathbb{P}(  Y_{0}- f(\bm{x}_0) \leq z)  |. \\
\end{split}
\end{equation}
For the first term on the r.h.s. of the above inequality, we have:
\begin{equation}\label{eq36}
\begin{split}
    &\sup_{z} | \mathbb{P}(Y^*_0 - f(\bm{x}_0)\leq z) - \mathbb{P}(Y_0^* - \overline{f}_{\text{DNN}}(\bm{x}_0)\leq z) | \\
    & = \sup_{z} | \mathbb{P}(Y_0^* - \overline{f}_{\text{DNN}}(\bm{x}_0) + \overline{f}_{\text{DNN}}(\bm{x}_0) -  f(\bm{x}_0)\leq z) - \mathbb{P}(Y_0^* - \overline{f}_{\text{DNN}}(\bm{x}_0)\leq z)\\
    &= \sup_{z}|F_{\epsilon_{0}^{*}}(z+f(\bm{x}_0) - \overline{f}_{\text{DNN}}(\bm{x}_0)) - F_{\epsilon_{0}^{*}}(z) |\\
    & = \sup_{z}| F_{\epsilon_{0}^{*}}(z+f(\bm{x}_0) - \overline{f}_{\text{DNN}}(\bm{x}_0)) - F_{\epsilon_{0}}(z+f(\bm{x}_0) - \overline{f}_{\text{DNN}}(\bm{x}_0)) \\
    &+ F_{\epsilon_{0}}(z+f(\bm{x}_0) - \overline{f}_{\text{DNN}}(\bm{x}_0)) - F_{\epsilon_{0}}(z) + F_{\epsilon_{0}}(z) -F_{\epsilon_{0}^{*}}(z)  |  \\
    & \leq \sup_{z}| F_{\epsilon_{0}^{*}}(z+f(\bm{x}_0) - \overline{f}_{\text{DNN}}(\bm{x}_0)) - F_{\epsilon_{0}}(z+f(\bm{x}_0) - \overline{f}_{\text{DNN}}(\bm{x}_0))| \\
    &+ \sup_{z}| F_{\epsilon_{0}}(z+f(\bm{x}_0) - \overline{f}_{\text{DNN}}(\bm{x}_0)) - F_{\epsilon_{0}}(z) | + \sup_{z}|F_{\epsilon_{0}}(z) -F_{\epsilon_{0}^{*}}(z)  |.
\end{split} 
\end{equation}
We should notice that the first and third terms of the r.h.s. of \cref{eq36} converge to 0 in probability. For the middle term, since $\overline{f}_{\text{DNN}}(\bm{x}_0)$ converges to $f(\bm{x}_0)$ in probability and $\sup_{z}|p_{\epsilon_{0}}(z)|$ is assumed to be bounded as B2, this term also converges to 0 in probability by applying the Taylor expansion. Combining all the pieces, we have:
\begin{equation}
    \sup_{z}\left| F_{Y_0^*|\bm{X}_0= \bm{x}_0}(z) -  F_{Y_{0}|\bm{X}_0= \bm{x}_0}(z) \right| \overset{p}{\to}0.
\end{equation}

\end{proof}

\section*{\textsc{Appendix B: Additional simulations on point estimations}}\label{Appendix:B}
In this part, we consider below models to check the performance of SS-DNN:
\begin{itemize}
    \item Model-1': $Y =  \sum_{i=1}^{10}X_i $, where $(X_1,\ldots, X_{10})\sim N(0,\bm{I})$. 
    \item Model-2': $Y = \sum_{i=1}^{10}i\cdot X_i$, where $(X_1,\ldots, X_{10})\sim N(0,\bm{I})$.
    \item Model-3': $Y = X_1^2 + \sin(X_2 + X_3)$, where $(X_1,X_2,X_3)\sim N(0,\bm{I})$. 
    \item Model-4': $Y = X_1^2 + \sin(X_2 + X_3) + \exp(-|X_4 + X_5|)$, where $(X_1,X_2,X_3,X_4,X_5)\sim N(0,\bm{I})$.
\end{itemize}

Similar to the main text, $\bm{I}$ is the identity matrix with the appropriate dimension. Compared to the data-generating model in the main text, the only difference is that the error term is removed. Due to this change, we can investigate the approximation ability of various DNN estimators straightforwardly, i.e., the MSE-1 error is equivalent to the MSE-2 error. Setting the same training procedure, we summarize all simulation results in \cref{Table: simresult}; here, empirical MSE and Run Times (in seconds) were also computed as averages of 200 replications.

\begin{table}[htbp]
  \centering
  \caption{MSE and Run Times of different DNN models with various simulation models}
  \vspace{2pt}
  
    \begin{tabular}{llcccccc}
    \toprule
        \multicolumn{2}{c}{} & \thead{ SS-DNN} & \thead{ S-DNN} & \thead{  DNN-deep-1 }& \thead{ DNN-deep-2 } & \thead{ DNN-wide-1  } & \thead{ DNN-wide-2 } \\
    \hline\\
    \multicolumn{2}{l}{Model-1 $n = 10^4$} &       &       &       &       &       &  \\[5pt]

    \multicolumn{2}{l}{Width} & [20,20] & [20,20] & [90,90] & [60,60] & [800] & [400] \\
    \multicolumn{2}{l}{MSE} & 0.0002 & 0.0008 & 0.0009 & 0.0007 & 0.0007 & 0.0006 \\
    \multicolumn{2}{l}{Run Times} & 238   & 252   & 446   & 338   & 405   & 298 \\
          &       &       &       &       &       &       &  \\
        \hline\\
    \multicolumn{2}{l}{Model-2 $n = 10^4$} &       &       &       &       &       &  \\[5pt]
    
    \multicolumn{2}{l}{Width} & [20,20] & [20,20] & [90,90] & [60,60] & [800] & [400] \\
    \multicolumn{2}{l}{MSE} & 0.0071 & 0.0126 & 0.0171 & 0.0213 & 0.0079 & 0.0104 \\
    \multicolumn{2}{l}{Run Times} & 209   & 227   & 406   & 308   & 367   & 270 \\
          &       &       &       &       &       &       &  \\
    \hline\\
    \multicolumn{2}{l}{Model-3 $n = 10^4$} &       &       &       &       &       &  \\[5pt]
 
    \multicolumn{2}{l}{Width} & [15,15,15] & [15,15,15] & [65,65,65] & [45,45,45] & [2000] & [1000] \\
    \multicolumn{2}{l}{MSE} & 0.0038 & 0.0075 & 0.0061 & 0.0065 & 0.0068 & 0.0066 \\
    \multicolumn{2}{l}{Run Times} & 251   & 271   & 445   & 356   & 397   & 289 \\

           &       &       &       &       &       &       &  \\
    \hline\\
    \multicolumn{2}{l}{Model-4 $n = 10^4$} &       &       &       &       &       &  \\[5pt]
 
    \multicolumn{2}{l}{Width} & [15,15,15] & [15,15,15] & [65,65,65] & [45,45,45] & [2000] & [1000] \\
    \multicolumn{2}{l}{MSE} & 0.0073 & 0.0126 & 0.0090 & 0.0096 & 0.0119 & 0.0124\\
    \multicolumn{2}{l}{Run Times} & 252   & 270   & 449   & 358   & 476   & 328 \\

               &       &       &       &       &       &       &  \\
    \hline\\
    \multicolumn{2}{l}{Model-4 $n = 2\cdot 10^4$} &       &       &       &       &       &  \\[5pt]
 
    \multicolumn{2}{l}{Width} & [20,20,20] & [20,20,20] & [95,95,95] & [65,65,65] & [2800] & [1400] \\
    \multicolumn{2}{l}{MSE} & 0.0083 & 0.0233 & 0.0194 & 0.0188 & 0.0247 & 0.0248 \\
    \multicolumn{2}{l}{Run Times} & 518   & 555   & 1438   & 962   & 1369   & 862 \\

    \bottomrule  
    \end{tabular}%
    \raggedright
    \vspace{2pt}
     \textit{Note:} Here, ``width'' represents the number of neurons of each hidden layer, e.g., [20, 20] means that there are two hidden layers within the DNN and each has 20 number neurons.
     \label{Table: simresult}
\end{table}%

The SS-DNN is still the most time-efficient estimator. It even runs faster than training S-DNN with the whole sample size. Applying the scalable subagging method can gain more computational savings for training with a larger sample size or a larger model. The SS-DNN is also the most accurate estimator except in the case with $10^4$ Model-4 simulated data. For this case, the accuracy of SS-DNN is slightly worse than the estimator DNN-deep-1. We conjecture the reason is that Model-4 is relatively complicated so a DNN with 3 depths and constant width 15 has a high bias. After increasing the sample size to 20000, the subagging estimator beats other models.

\section*{\textsc{Appendix C: Simulations for unconditional CI and PI}}\label{Appendix:C}
In this part, we consider the unconditional (not conditional on $\bm{X}_0 = \bm{x}_0$) performance of CI and PI defined in the main text. We take the below empirical coverage rate (ECR) and empirical length of CI (EL) as the measurement criteria for (unconditional) CI:
\begin{equation*}
     \text{ECR}= \frac{1}{n}\sum_{k=1}^{n}\mathbbm{1}_{f(\bm{x}_{k})\in [B_{l,k},B_{u,k}]}~;~\text{EL}= \frac{1}{n}\sum_{k=1}^{n} (B_{u,k} -  B_{l,k});
\end{equation*}
here $f(\bm{x}_{k})$ is the true model value at the $k$-th data point in the training dataset; $B_{u,k}$ and $B_{l,k}$ are the corresponding upper and lower bounds of CI, respectively. To measure the performance of (unconditional) PI, we take 
$$\text{ECR}= \frac{1}{N}\sum_{i=1}^{N}\mathbbm{1}_{y_{i,0}\in [B_{l,i},B_{u,i}]}~,~\text{EL}= \frac{1}{N}\sum_{i=1}^{N} (B_{u,i} -  B_{l,i});$$
here $y_{i,0}$ is the $i$-th observed response value in the test dataset; $B_{u,k}$ and $B_{l,k}$ are the corresponding upper and lower bounds of PI. We take $n = N = 2\cdot 10^5$. Due to the (unconditional) ECR and EL hardly changing in the simulation studies, we just do 50 replications and we present the average results of various CIs and PIs in \cref{Table: simresulterrorCI} and \cref{Table: simresulterrorPI}, respectively. Similar to simulation results of conditional CIs and PIs in the main text, QCI-1 undercovers true model values; PCI-2 shows the best comprehensive performance according to length and coverage rate. All PIs show great performance but slightly undercover true future values.

For the computational issue of the iterated subsampling stage, the total time of training all DNN estimators $\widehat{f}^{(j)}_{\text{DNN},b,i}$ for $i \in\{1,\ldots,q\}$ and $j\in\{1,\ldots,q^{\prime} \}$ (iterated subsampling stage) is less than the time of training all DNN estimators in the first subsampling stage, i.e., $\widehat{f}_{\text{DNN},b,i}$ for $i \in\{1,\ldots,q\}$. We can see the reason by analyzing the computational complexity of the iterated subsampling stage. In total, we need to train $q\cdot q^{\prime} = O(n^{1-\beta^2})$ number of models with sample size $n^{\beta^2}$. As the assumption we made in \cref{Remark:complexityabalysis}, the complexity of training a DNN is mainly determined by its size, sample size and the number of epochs, so the training time of the iterated stage is around $q\cdot q^{\prime} \cdot O(n^{\beta^2}\cdot n^{\beta^2}) = O(n^{1 + \beta^2  })$ when the sample size is close to the size of DNN. Similarly, we can analyze that the complexity of training DNNs in the first subsampling stage is around $O(n^{1 + \beta })$. Since $\beta<1$, the complexity of the first subsampling stage will dominate the iterated stage when $n$ is large enough. In other words, the complexity cost of applying the iterated subsampling technique is negligible when we are dealing with a huge dataset.

\begin{table}[htbp]

  \caption{Empirical Coverage Rate and Empirical Length of different (unconditional) CIs with various simulation models; Run Times of first and iterated subsampling stages}
  \vspace{2pt}
  
    \begin{tabular}{llccccc}
    \toprule
        \multicolumn{2}{c}{Nominal $\delta$: } & & & \thead{0.1} & \thead{0.05} \\
    \hline
    \multicolumn{3}{l}{Model-1, $n = 2\cdot 10^5$  } &   & &               \\[3pt]

    \multicolumn{2}{l}{ECR of QCI-1} & & & 1.000 & 1.000 \\
    \multicolumn{2}{l}{ECR of QCI-2} & & & 0.961 & 0.983 \\
    \multicolumn{2}{l}{ECR of PCI-1} & & & 0.842 & 0.905  \\
    \multicolumn{2}{l}{ECR of PCI-2} & & & 0.953 & 0.976 \\
    \multicolumn{2}{l}{ECR of PCI-3} & & & 0.853 & 0.914\\
    \multicolumn{2}{l}{EL of QCI-1} & & & 1.752 & 2.371 \\
    \multicolumn{2}{l}{EL of QCI-2} & & & 1.090 & 1.316 \\
    \multicolumn{2}{l}{EL of PCI-1} & & & 0.292 & 0.348  \\
    \multicolumn{2}{l}{EL of PCI-2} & & & 0.549 & 0.654   \\
    \multicolumn{2}{l}{EL of PCI-3} & & & 0.299 & 0.356   \\
     \multicolumn{3}{l}{Training time of first and iterated subagging estimators:} &   \multicolumn{3}{l}{5613 $\&$ 4566}         \\
      \multicolumn{3}{l}{Structure of $\widehat{f}_{\text{DNN},b,i}$ and $\widehat{f}^{(j)}_{\text{DNN},b,i}$:} & \multicolumn{3}{l}{[65,65] $\&$ [10,10]}         \\[5pt]

    \multicolumn{3}{l}{Model-2, $n = 2\cdot 10^5$  } &   & &    &           \\[3pt]

    \multicolumn{2}{l}{ECR of QCI-1} & & & 1.000 & 1.000 \\
    \multicolumn{2}{l}{ECR of QCI-2} & & & 0.925 & 0.961  \\
    \multicolumn{2}{l}{ECR of PCI-1} & & &  0.875 & 0.931  \\
    \multicolumn{2}{l}{ECR of PCI-2} & & & 0.957 & 0.979 \\
    \multicolumn{2}{l}{ECR of PCI-3} & & & 0.883 & 0.937  \\
    
    \multicolumn{2}{l}{EL of QCI-1} & & & 2.213 & 2.847 \\
    \multicolumn{2}{l}{EL of QCI-2} & & & 1.117 & 1.349  \\
    \multicolumn{2}{l}{EL of PCI-1} & & & 0.359 & 0.428  \\
    \multicolumn{2}{l}{EL of PCI-2} & & & 0.597 & 0.711  \\
     \multicolumn{2}{l}{EL of PCI-3} & & & 0.367 & 0.437  \\
     \multicolumn{3}{l}{Training time of first and iterated subagging estimators:} &   \multicolumn{3}{l}{5451 $\&$ 4566}         \\ 
      \multicolumn{3}{l}{Structure of $\widehat{f}_{\text{DNN},b,i}$ and $\widehat{f}^{(j)}_{\text{DNN},b,i}$:} & \multicolumn{3}{l}{[65,65] $\&$ [10,10]}         \\[5pt]

    \multicolumn{3}{l}{Model-3, $n = 2\cdot 10^5$  } &   & &              \\[3pt]

    \multicolumn{2}{l}{ECR of QCI-1} & & & 1.000 & 1.000   \\
    \multicolumn{2}{l}{ECR of QCI-2} & & & 0.967 & 0.985 \\
    \multicolumn{2}{l}{ECR of PCI-1} & & & 0.776 & 0.846  \\
    \multicolumn{2}{l}{ECR of PCI-2} & & & 0.961 & 0.976 \\
    \multicolumn{2}{l}{ECR of PCI-3} & & & 0.802 & 0.867 \\
    
    \multicolumn{2}{l}{EL of QCI-1} & & & 0.774 & 0.956  \\
    \multicolumn{2}{l}{EL of QCI-2} & & & 0.625 & 0.755 \\
    \multicolumn{2}{l}{EL of PCI-1} & & & 0.125 & 0.149 \\
    \multicolumn{2}{l}{EL of PCI-2} & & & 0.457 & 0.545  \\
    \multicolumn{2}{l}{EL of PCI-3} & & & 0.132 & 0.158  \\
     \multicolumn{3}{l}{Training time of first and iterated subagging estimators:} &   \multicolumn{3}{l}{6597 $\&$ 5551 }      \\ 
      \multicolumn{3}{l}{Structure of $\widehat{f}_{\text{DNN},b,i}$ and $\widehat{f}^{(j)}_{\text{DNN},b,i}$:} & \multicolumn{3}{l}{[45,45,45] $\&$ [10,10,10]}         \\[5pt]
     
         \multicolumn{3}{l}{Model-4, $n = 2\cdot 10^5$  } &   & &             \\[3pt]

    \multicolumn{2}{l}{ECR of QCI-1} & & & 1.000 & 1.000  \\
    \multicolumn{2}{l}{ECR of QCI-2} & & & 0.904 & 0.945 \\
    \multicolumn{2}{l}{ECR of PCI-1} & & & 0.762 & 0.830  \\
    \multicolumn{2}{l}{ECR of PCI-2} & & & 0.931 & 0.956 \\
    \multicolumn{2}{l}{ECR of PCI-3} & & & 0.778 & 0.844 \\
    
    \multicolumn{2}{l}{EL of QCI-1} & & &  1.242 & 1.570 \\
    \multicolumn{2}{l}{EL of QCI-2} & & & 0.903 & 1.091 \\
    \multicolumn{2}{l}{EL of PCI-1} & & &  0.202 & 0.240 \\
    \multicolumn{2}{l}{EL of PCI-2} & & & 0.496 & 0.591\\
    \multicolumn{2}{l}{EL of PCI-3} & & & 0.209 & 0.249  \\
     \multicolumn{3}{l}{Training time of first and iterated subagging estimators:} &   \multicolumn{3}{l}{6451 $\&$ 5419 }   \\ 
          \multicolumn{3}{l}{Structure of $\widehat{f}_{\text{DNN},b,i}$ and $\widehat{f}^{(j)}_{\text{DNN},b,i}$:} & \multicolumn{3}{l}{[45,45,45] $\&$ [10,10,10]}         \\
     
    \bottomrule  
    \end{tabular}%
    %\raggedright
    %\vspace{2pt}
    
     %\textit{Note:} The specified structure of $\widehat{f}_{\text{DNN},b,i}$ and $\widehat{f}^{(j)}_{\text{DNN},b,i}$ works for any $i \in\{1,\ldots,q\}$ and $j\in\{1,\ldots,q^{\prime} \}$.
     \label{Table: simresulterrorCI}
\end{table}%

\FloatBarrier

\begin{table}[htbp]
\centering
  \caption{Empirical Coverage Rate and Empirical Length of (unconditional) PI with various simulation models}
  \vspace{2pt}
  
    \begin{tabular}{llccccc}
    \toprule
        \multicolumn{2}{c}{Nominal $\delta$: } & & & \thead{0.1} & \thead{0.05} \\
    \hline
    \multicolumn{3}{l}{Model-1, $N = 2\cdot 10^5$  } &   & &               \\[3pt]

    \multicolumn{2}{l}{ECR of PI} & & & 0.8981 & 0.9486 \\
    \multicolumn{2}{l}{EL of PI} & & & 3.2919 & 3.9234  \\[5pt]

    \multicolumn{3}{l}{Model-2, $N = 2\cdot 10^5$  } &   & &    &           \\[3pt]

    \multicolumn{2}{l}{ECR of PI} & & & 0.8983 &  0.9487 \\
    \multicolumn{2}{l}{EL of PI} & & & 3.3008 & 3.9316 \\ [5pt]

    \multicolumn{3}{l}{Model-3, $N = 2\cdot 10^5$  } &   & &              \\[3pt]

    \multicolumn{2}{l}{ECR of PI} & & & 0.8973 &  0.9481   \\
    \multicolumn{2}{l}{EL of PI} & & & 3.3022 & 3.9339  \\  [5pt]

         \multicolumn{3}{l}{Model-4, $N = 2\cdot 10^5$  } &   & &             \\[3pt]

    \multicolumn{2}{l}{ECR of PI} & & & 0.8975 & 0.9482  \\
    \multicolumn{2}{l}{EL of PI} & & & 3.3135 & 3.9411  \\

    \bottomrule  
    \end{tabular}%
    %\raggedright
    %\vspace{2pt}
    
     %\textit{Note:} The specified structure of $\widehat{f}_{\text{DNN},b,i}$ and $\widehat{f}^{(j)}_{\text{DNN},b,i}$ works for any $i \in\{1,\ldots,q\}$ and $j\in\{1,\ldots,q^{\prime} \}$.
     \label{Table: simresulterrorPI}
\end{table}%

\clearpage
\bibliographystyle{apalike}
\bibliography{refs}
\end{document}